\pdfoutput=1
\PassOptionsToPackage{dvipsnames,svgnames,x11names}{xcolor}
\documentclass[11pt]{article}

\usepackage{acl}

\usepackage{times}
\usepackage{latexsym}

\usepackage[T1]{fontenc}
\usepackage[utf8]{inputenc}

\usepackage{microtype}

\usepackage{inconsolata}

\usepackage{times}
\usepackage{latexsym}
\usepackage[T1]{fontenc}
\usepackage[utf8]{inputenc}
\usepackage{microtype}
\usepackage{inconsolata}
\usepackage{dutchcal}
\usepackage{algorithm, algorithmicx, algpseudocode}

\usepackage{multirow}
\usepackage{graphicx}
\usepackage{booktabs}
\usepackage{amsthm,amsmath,amssymb}
\usepackage{bm}
\usepackage[normalem]{ulem}
\useunder{\uline}{\ul}{}

\usepackage{subfigure}
\usepackage{tabularx}
\usepackage{xspace}

\usepackage{enumitem}
\setenumerate[1]{itemsep=0pt,parsep=0pt,topsep=0pt}
\setitemize[1]{itemsep=5pt,parsep=0pt,topsep=5pt,leftmargin=8pt}
\newcommand{\sysname}{\textbf{\texttt{CARE}}\xspace}
\title{\sysname : A Clue-guided Assistant for CSRs to Read User Manuals}

\author{
{Weihong Du\textsuperscript{1}\textsuperscript{2}\quad Jia Liu\textsuperscript{1}\quad Zujie Wen\textsuperscript{3}}
\\
{\textbf{Dingnan Jin}\textsuperscript{4}\quad 
\textbf{Hongru Liang}\textsuperscript{1}\textsuperscript{2}\thanks{\quad Corresponding author}\quad
\textbf{Wenqiang Lei}\textsuperscript{1}\textsuperscript{2}}
 \\
{\textsuperscript{1}College of Computer Science, Sichuan University, China} \\
{\textsuperscript{2}Engineering Research Center of Machine Learning and Industry Intelligence,} \\
{Ministry of Education, China} \\
{\textsuperscript{3}Dalian University of Technology, China}\\
{\textsuperscript{4}University of Electronic Science and Technology, China} \\
\texttt{duweihong@stu.scu.edu.cn} \quad
\texttt{\{lj\_m, wenzj2010\}@163.com} \quad \\
\texttt{alex2triten@gmail.com}\\
\texttt{\{lianghongru, wenqianglei\}@scu.edu.cn}\\ 
}

\begin{document}

\maketitle

\begin{abstract}
It is time-saving to build a reading assistant for customer service representations~(CSRs) when reading user manuals, especially information-rich ones. Current solutions don't fit the online custom service scenarios well due to the lack of attention to user questions and possible responses. Hence, we propose to develop a time-saving and careful reading assistant for CSRs, named \sysname. It can help the CSRs quickly find proper responses from the user manuals via explicit clue chains. Specifically, each of the clue chains is formed by inferring over the user manuals, starting from the question clue aligned with the user question and ending at a possible response. To overcome the shortage of supervised data, we adopt the self-supervised strategy for model learning. The offline experiment shows that \sysname is efficient in automatically inferring accurate responses from the user manual. The online experiment further demonstrates the superiority of \sysname to reduce CSRs' reading burden and keep high service quality, in particular with $>35\%$ decrease in time spent and keeping a $>0.75$ ICC score. \sysname is available in \url{https://github.com/SCUNLP/CARE}. 
\end{abstract}

\begin{figure*}[!t]
    \centering
    \includegraphics[width=0.96\textwidth]{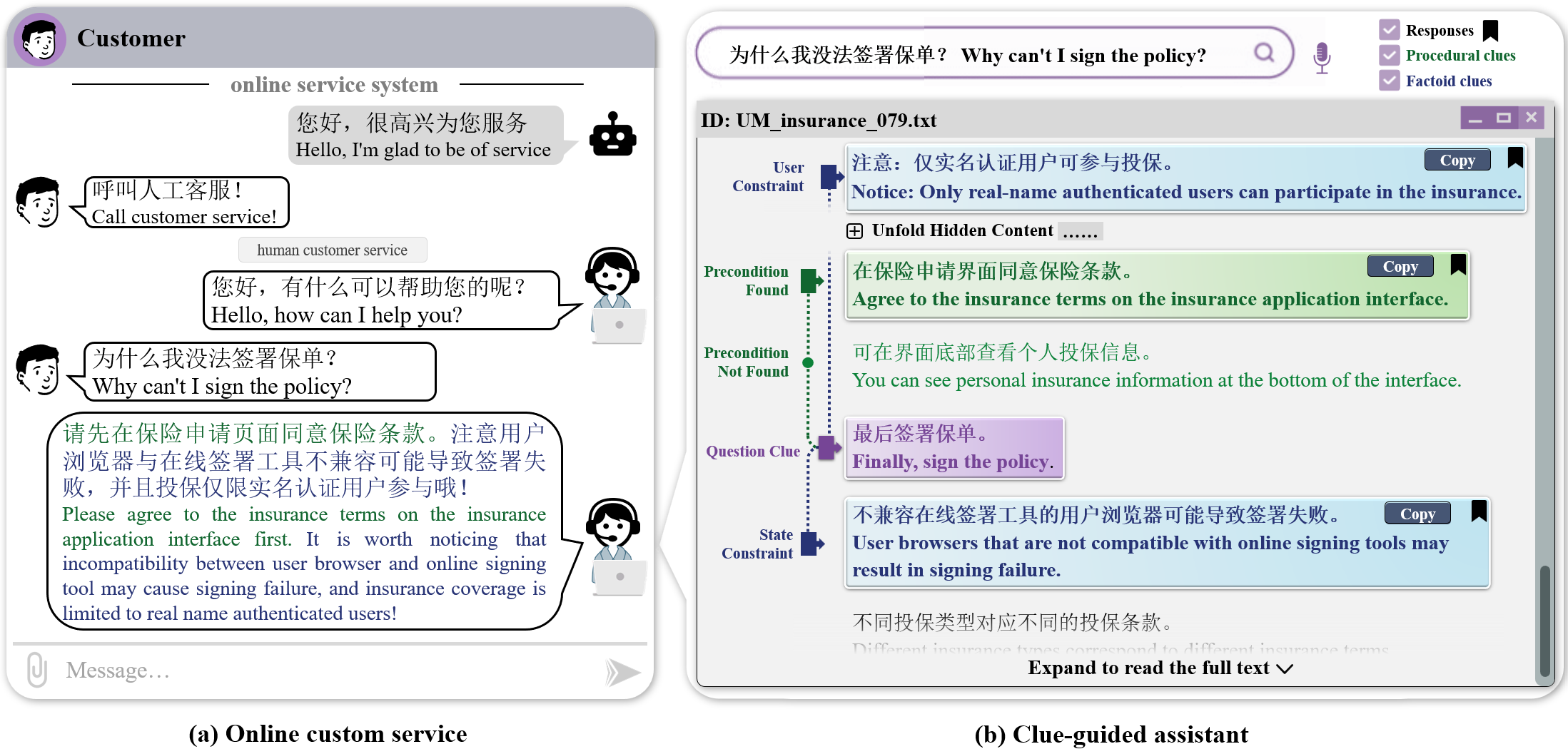}
    \caption{Illustrations of (a) a log of online custom service, where the user prefers to chat with CSR; and (b) the proposed clue-guided reading assistant~(\sysname), which provides clues explaining how to arrive at proper responses.}
    \label{fig:assistance}
\end{figure*}
\section{Introduction}
\label{sec:intro}

In e-commerce, human services are always involved --- once called, a custom service representation~(CSR) is required to answer the user's questions based on a related user manual~\cite{article}. This job is not easy, as the CSR needs to read the information-rich user manual and compose a proper response to the user, {as shown in Figure~\ref{fig:assistance}~(a)}. It will be greatly time-saving if we build a reading assistant for CSRs when reading user manuals and seeking possible responses.\par

Current reading assistants~\cite{badam2018elastic, fok2023scim} have demonstrated that highlighting special contents~(e.g., the co-reference phrases in a document, the ``Notice'' sections of a document, etc) can reduce people's reading burden when skimming over a whole document. However, these solutions don't work well in online customer services because the highlighted contents have few connections to the question raised by a user or possible responses to the question. There is still a blank of assistants that virtually contribute to saving CSRs' time in reading user manuals for locating proper responses to users quickly. \par

Hence, we consider building a reading assistant for CSRs that can highlight possible responses, which are expected to be accurate enough to greatly save CSRs' time by 
% so that the CSRs can save a lot of time by 
copying these responses. Beyond this, we are concerned about its practicability in real-world scenarios --- if a CSR accidentally copies the responses wrongly predicted by the assistant, it will risk the service quality of online custom service. This forces us to build a more careful reading assistant that can also explicitly explain how to arrive at these responses from the user question. As a remedy, we propose to explain responses via clue chains, cf., Figure~\ref{fig:assistance} (b). Each of the clue chains starts from the question clue, walks through transitional clues, and ends at a response clue. The transitional clues explain how the response clues connect with the question clue. For example, the transitional clue ``\textcolor{DarkGreen}{You can see ... the interface}'' indicates that the response clue ``\textcolor{DarkGreen}{Agree to the ... application interface}'' is two steps before ``\textcolor{DarkOrchid}{Finally, sign the policy}''. Note that, besides steps, the facts binding the steps may also be response clues, e.g., ``\textcolor{DarkBlue}{Notice: Only real-name ... the insurance}''. As such, the CSRs can check the correctness of response clues more easily and are less likely to copy improper responses predicted by the model. \par

We call this \underline{C}lue-guided \underline{A}ssistant for CSRs to \underline{RE}ad user manuals as \sysname. 
To get explainable clue chains over unstructured user manuals, we convert them into heterogeneous graphs. As such, the explainable comprehension of user manuals turns to explicit inferring over the graphs. Accordingly, the alignment of the user question and its question clue in the user manual can be modeled as finding the most relevant node to the user question in the graph. This is done with the aid of a matrix projecting the user question to the graph space, which is founded on the heterogeneous graph attention network. After getting the question clue node, we model the inference of response clues as the link prediction task from the question clue node to proper response clues. Further, to overcome the shortage of dedicated supervision data, we adopt the self-supervised training strategy to enhance the reasoning ability of the model from the inherent information of user manuals. The offline experiment on the testing set, derived from a Chinese e-commerce platform, demonstrates the efficiency of \sysname to infer accurate responses to questions. Besides, the online experiment, deployed to train new CSRs in practice, demonstrates that \sysname can significantly reduce the reading burden of CSRs~(saving more than $35\%$ time) and maintain high service quality~($>0.75$ ICC score). We summarize our contributions as follows.

\begin{itemize}
    \item We call attention to the importance of building a time-saving and careful reading assistant in online custom service. It must be efficient to provide accurate responses and be explainable to guide the CSRs to arrive at these responses.
    \item We propose to use clue chains to achieve such an assistant and develop \sysname, which can not only obtain proper responses to users but also explain the responses by inferring over structured representations of user manuals.
    \item Both offline and online experiments demonstrate the superiority of \sysname in reducing CSR's reading burden and keeping high service quality. 
\end{itemize}

\section{Related Work}
\paragraph{Reading Assistant} 
Designing an efficient assistant to reduce the human reading burden is vital in building an enjoyable reading experience for information-rich documents~\cite{badam2018elastic}. 
% Most of the current solutions have well-demonstrated that highlighting special contents from the documents can promote human reading speed and document reading experience. 
A line of work claims that connecting co-reference contents can simplify the reading process~\cite{strobelt2009document, badam2018elastic, pinheiro2022charttext}, for example, highlighting all words related to ``insurance'' in the user manual of Figure~\ref{fig:assistance}. Another line of work focuses on marking pre-defined and specific contents in the documents~\cite{yang2017hitext, cachola2020tldr, fok2023scim}, for example, highlighting all ``Notice'' sections in the user manual. Although these solutions can direct the reader's focus to the main idea of a document, they can hardly reduce the CSRs' burden because these highlighted contents have few connections to the user question and possible responses. 
% A simple alternative is to highlight keywords of the question~(e.g., ``{enter the password}'' of the question in Figure~\ref{fig:assistance}) in the user manual. 
% However, the responses are not always closely surrounding the keywords of questions~\cite{dalvi2018tracking}, cf., Figure~\ref{fig:assistance} (b).
To fill this blank, we develop a clue-guided assistant that saves CSRs' time by explicitly explaining how to get accurate responses in the user manual.\par
\begin{figure*}[t]
    \centering
    \includegraphics[width=0.9\textwidth]{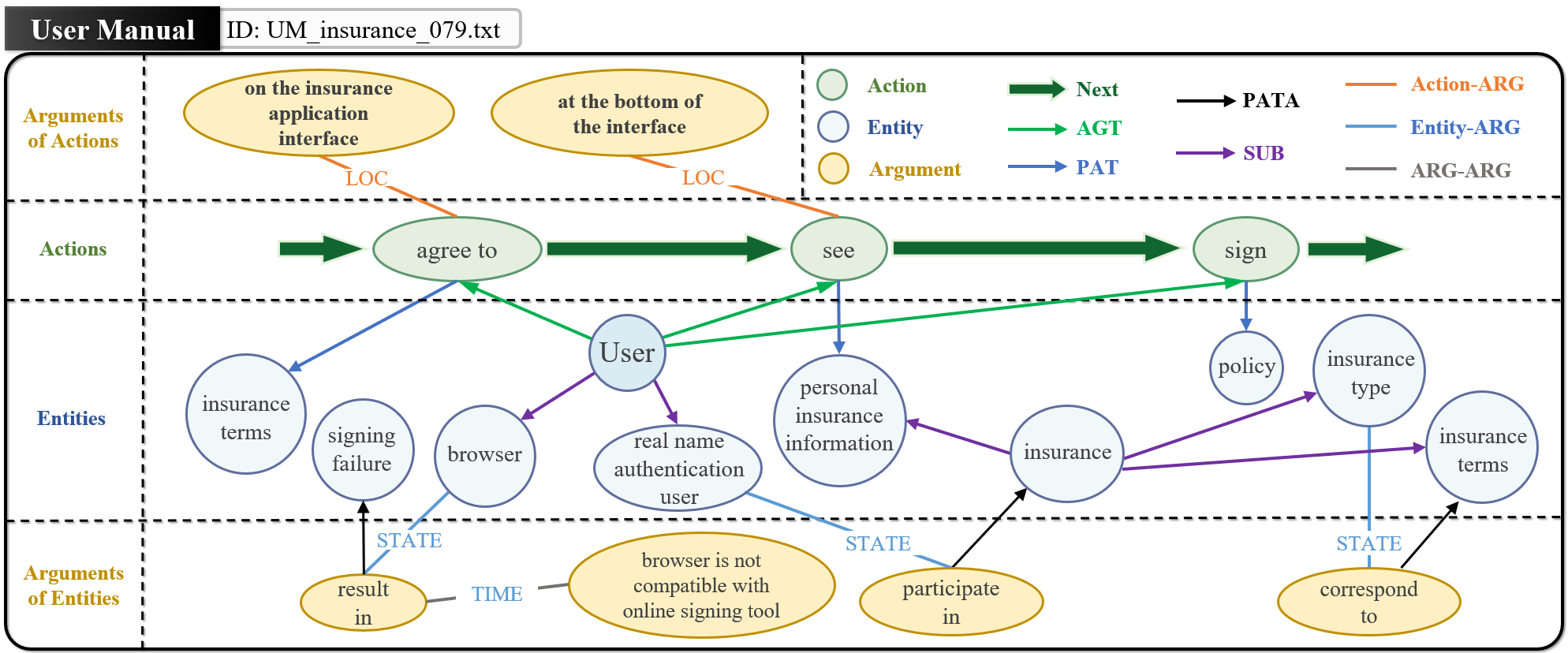} 
    \caption{Part of the heterogeneous graph derived from the user manual in Figure~\ref{fig:assistance}. Each action on the graph is decorated with corresponding arguments, and the arguments possessed by each entity and the sub-entity relations between different entities are clearly represented by corresponding links. Best viewed in color.}
    \label{fig:graph}
     \vspace*{-0cm}
\end{figure*}
\paragraph{Explainable Machine Comprehension} Despite growing interest, explainable machine comprehension, which requires the model to expose how to arrive at the final responses/answers given a document, is still in the primary stage~\cite{thayaparan2020survey}. Therefore, most of the existing works are devoted to constructing explanation-supporting benchmarks in machine comprehension~\cite{khashabi2018looking,inoue2020r4c,ho2020constructing,cui2022expmrc}. However, these annotated explanations only contribute to factoid-style reasoning~\cite{li2021asynchronous,tan2021gcrc,zhao2023hop}.
There is still a lack of explanation-supporting datasets for user manuals, which contain both steps and facts. This motivated us to leverage the self-information in user manuals and design a self-supervised strategy for model learning. 
Specifically, we propose an innovative data construction approach that generates both factual and procedural Q\&A pairs from user manuals. These augmented data can greatly enhance the model's ability to answer complex user questions.

\paragraph{Knowledge Graph Question Answering} Compared with conventional question answering, knowledge graph question answering has the inherent advantage of forming explanations~(i.e., inferring path) leading to the answers~\cite{yasunaga2021qa,zhang2022fact}.
% ,lan2022complex}. 
This inspires us to convert raw user manuals into structured graphs. In this way, the comprehension of user manuals becomes explicit inferring over the nodes and edges of the graphs. 
However, current works only characterize factual knowledge in the knowledge graph~\cite{talmor2018web, lan2021survey},
% , yasunaga2021qa}, 
overlooking the necessity of procedural knowledge for integrated understanding of user manuals. As a remedy, we represent the user manuals as heterogeneous graphs to support the reasoning of responses to complex user questions with the help of both procedural and factual information. 

\paragraph{Heterogeneous Graph Reasoning} Reasoning on heterogeneous graphs, which are organized as multiple types of nodes and edges, has been used in various tasks that require inference on complex information~\cite{Wang2019HeterogeneousGA, bing2023heterogeneous, wang2023survey}. For example, \citet{wang2021self} and \citet{yu2023heterogeneous} proposed cross-view contrastive learning based reasoning algorithms to conduct in-depth reasoning for node classification, node clustering, graph classification, etc. In line with these works, we believe the heterogeneous graph reasoning techniques will also greatly benefit the deep comprehension of a user manual, which contains both factual and procedural types of information. 
Despite the big potential, similar efforts are seldom seen in handling user manuals, which also contain multiple types of information. An exception is \citet{liang2023knowing}, which is the only surveyed literature representing user manuals as heterogeneous graphs. However, this work uses manually-defined rules to find potential answers to user questions and thus has trouble answering questions beyond the rules. To this end, in this paper, we contribute the primary attempt to learn a heterogeneous graph reasoning model for user manuals and cope with various user questions through explicit inference on both procedural and factual information over graphs. 

% Although heterogeneous graphs have been widely used to encode different semantic relationships by designing various meta-paths~\cite{fu2020magnn, wang2021self, yu2023heterogeneous}, we are the first to employ heterogeneous graphs to jointly represent procedural and factual information in user manuals. 
% To better support the reasoning of complex user questions, we proposed two kinds of meta-paths to simultaneously 
% represent the procedural and factual information in user manuals and adopt them for joint inference. 

\section{Methodology}

\begin{figure*}[t]
    \centering
    \includegraphics[width=0.95\textwidth]{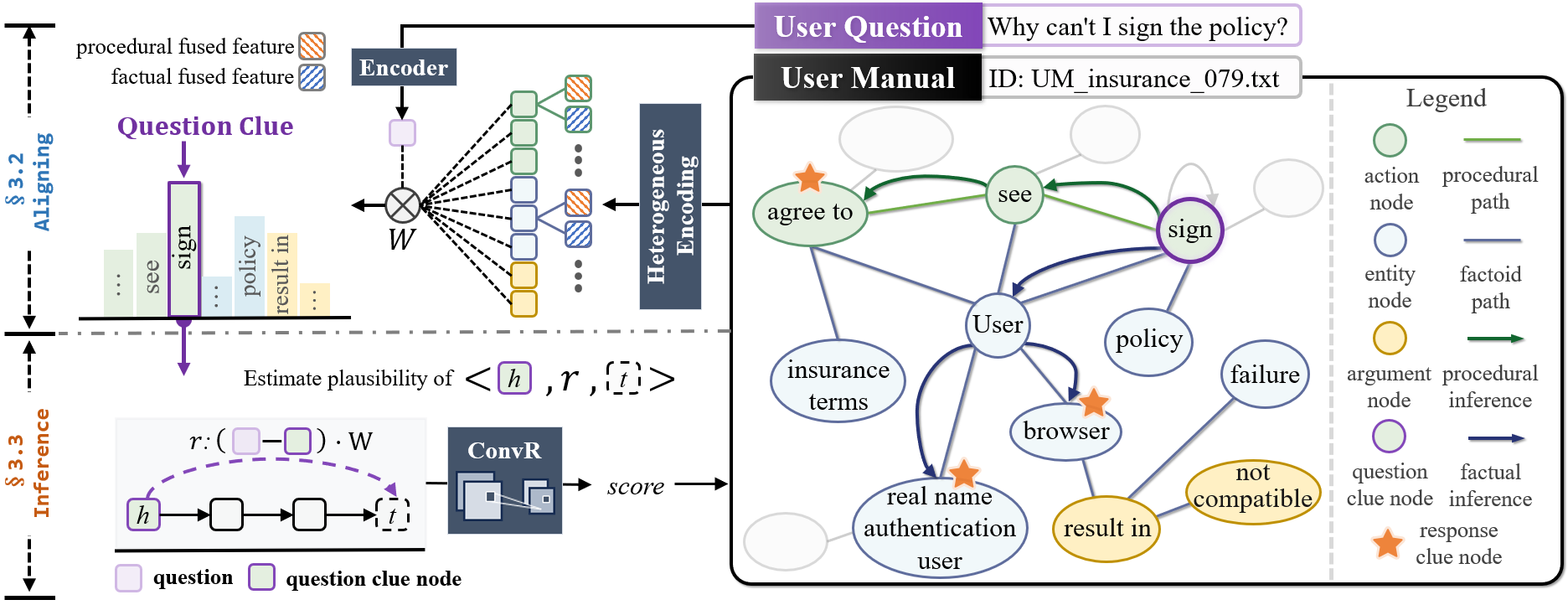}
    \caption{The backbone model of \sysname, which first aligns the user question to the graph of the user manual and then infers procedural and factoid clues over the graph from the question clue node.}
    \label{fig:framework}
\end{figure*}

The proposed reading assistant \sysname is founded on accurate responses to user questions and explicit explanations of these responses. We model it as inferring over the structured representations~(i.e., heterogeneous graphs, as shown in Figure~\ref{fig:graph}\footnote{For convenience and saving spaces, some examples are only presented in English.}) of user manuals given user questions. 
Specifically, we propose a backbone model to infer responses as illustrated in Figure~\ref{fig:framework}. 
The user question is aligned to the user manual by finding the most related node in the graph, namely the question clue node. Starting from the question clue node, we jointly infer procedural and factual clues over the graph until reach the response clue nodes that can compose proper responses to the user question.

\subsection{Heterogeneous Graph Construction}
\label{graphs detail}
Inspired by \citet{liang2023knowing}, we represent the user manual as a heterogeneous graph, which consists of action nodes, entity nodes, arguments nodes, and their relations, as shown in Figure~\ref{fig:graph}. \par
% Here we provide a detailed introduction to the construction of the heterogeneous graph of the user manual. 
% As preparation, we first use the off-the-shelf parsing tool to identify actions, entities, and their corresponding arguments from the user manuals. 
The procedures of a user manual are formed as several sequences of action nodes by linking actions in the same procedure with the ``Next'' relation. For example, the relation from the action node ``see'' to the action node ``sign'' is ``Next''.
% We use the Next Sentence Prediction(NSP) of the pre-trained model~\cite{shi2019next} to determine whether two actions belong to the same procedure and construct all the procedures. 
% To establish connections between agents, patients, and actions, 
The ``AGT'' relation links an action node with its agent~(e.g., ``User'') and the ``PAT'' relation connects each patient~(e.g., ``policy'') with their action. The ``Action-ARG'' relation links an action node with its arguments, including the time, location, and manner descriptions. \par
Entity nodes are derived from the concepts\footnote{Each user manual has a ``User'' entity by default.} in the user manual and linked with their arguments via the ``Entity-ARG'' relations. For example, the state node ``result in'' is a changeable state argument of the entity ``browser''. We allow an entity to have sub-entities and link them via the ``SUB'' relation. The ``PATA'' relation links the state node to an entity node, denoting that the entity is affected by the changing of state. The ``ARG-ARG'' relation describes the augments of an entity's argument, e.g., the time argument of the state argument ``result in''.

% We also link the states of the entity with corresponding entities affected by it through the ``PATA'' relation and link the arguments associated with other entities' arguments through the ``ARG-ARG'' relation. 

% Finally, We adopt character level matching to confirm the same entities and arguments references and fuse the same reference nodes into one node to avoid duplicate reference nodes of the constructed graphs. 
% Specifically, ``USER'' denotes the user entity node, which is created by default for each user manual. This user entity node describes all user constraints of the user manual. 

% In summary, the constructed heterogeneous graph consists of three kinds of nodes (i.e. action nodes, entity nodes, and argument nodes) and eight kinds of relations (i.e. Next, AGT, PAT, PATA, SUB, Action-ARG, Entity-ARG, and ARG-ARG). 
% To better support the heterogeneous inference of our proposed model on the graph, we reduce the complexity of the graph structure through retaining the original three node types (i.e. action nodes, entity nodes, and argument nodes) and classifying edge types into procedural and factual links. The procedural link only includes the ``Next'' relation, while the remaining relations belong to the factual link. 

Notably, towards joint inference of procedural and factual knowledge, we summarize the above-mentioned heterogeneous graph into three types of nodes~(i.e., action nodes, entity nodes, and argument nodes) and two types of paths~((i.e., the procedural path and the factual path). The procedural path starts from an action node
and walks over the graph directed by the ``Next'' relation. Besides the starting node, all nodes passed by the procedural path are action nodes. What's more, a path constructed from the remaining relations is defined as a factual path.

% We retain the original three node types (i.e. action nodes, entity nodes, and argument nodes) and classifying edge types into procedural and factual links. The procedural link only includes the ``Next'' relation, while the remaining relations belong to the factual link. 

\subsection{Alignment}
% To jointly infer procedural and factoid clues from user manuals, we represent the user manual as a heterogeneous graph~\cite{liang2023knowing}, which consists of action nodes, entity nodes, arguments nodes, and their relations, as shown in the right part of Figure~\ref{fig:framework}. Specifically, ``USER'' denotes the user entity node, which is created by default for each user manual. This user entity node describes all user constraints of the user manual. 
As illustrated in Figure~\ref{fig:framework}, the first part of the backbone model is aligning the user question to a node in the graph $G = \{ V, E \}$, namely the question clue node. Following~\cite{Wang2019HeterogeneousGA}, we adopt the heterogeneous graph attention network to encode the action, entity, and argument nodes into vectors. \par
% Specifically, we summarize the edges into two types of meta-paths in the graph --- the procedural paths refer to the edge between action nodes and the factoid paths refer to the remaining edges between nodes. 
Specifically, the procedural neighboring nodes set of a node is collected from 1-hop procedural paths, and the factual neighboring nodes set is collected from 1-hop factual paths. 
The embedding of a node is obtained from two-step fusion --- the embeddings of all nodes in the same neighboring nodes set are fused to produce a set embedding; all set embeddings are fused to produce the embedding of the current node through path-level attention. This process can be formulated as
% \vspace*{-0.3cm}
% \begin{equation}
% \small
%     Z = Weight_{\mathbb{P}} \cdot Z_{\mathbb{P}} + Weight_{\mathbb{F}} \cdot Z_{\mathbb{F}}
% \end{equation}

% \begin{equation}
% \small
%     ( Weight_{\mathbb{P}}, Weight_{\mathbb{F}}) = Attention(Z_{\mathbb{P}}, Z_{\mathbb{F}})
% \end{equation}
\vspace*{-1em}
\begin{equation}
\footnotesize
% 2^{\tt nd}\texttt{-step fusion}:
    \mathtt{\mathbf{n}} = \alpha_{\mathbb{P}} \mathtt{\mathbf{s}}_{\mathbb{P}}+\alpha_{\mathbb{F}}\mathtt{\mathbf{s}}_{\mathbb{F}},
    % \vspace*{-0.6em}
\end{equation}
% \begin{equation}
% \footnotesize
%     % \texttt{weight}: 
%     \alpha_{\mathbb{P}}={\tt softmax \big(\frac{\sum \mathtt{\mathbf{s}}_{\mathbb{P}}}{\sum \mathtt{\mathbf{s}}_{\mathbb{P}}+\sum \mathtt{\mathbf{s}}_{\mathbb{F}}}\big)},
%     \vspace*{-0.6em}
% \end{equation}
\begin{equation}
\footnotesize
    % \texttt{weight}: 
    \alpha_{\mathbb{P}}={\tt softmax \big[({\sum \mathtt{\mathbf{s}}_{\mathbb{P}}}\big/({\sum \mathtt{\mathbf{s}}_{\mathbb{P}}+\sum \mathtt{\mathbf{s}}_{\mathbb{F}}})\big]},
    % \vspace*{-0.6em}
\end{equation}
% \begin{equation}
% \footnotesize
%     \alpha_{\mathbb{F}}={\tt softmax \big(\frac{\sum \mathtt{\mathbf{s}}_{\mathbb{F}}}{\sum \mathtt{\mathbf{s}}_{\mathbb{P}}+\sum \mathtt{\mathbf{s}}_{\mathbb{F}}}\big)},
%     \vspace*{-0.6em}
% \end{equation}
\begin{equation}
\footnotesize
    \alpha_{\mathbb{F}}={\tt softmax \big[{\sum \mathtt{\mathbf{s}}_{\mathbb{F}}}\big/({\sum \mathtt{\mathbf{s}}_{\mathbb{P}}+\sum \mathtt{\mathbf{s}}_{\mathbb{F}}})\big]},
    % \vspace*{-0.6em}
\end{equation}
%  
% \begin{equation}
% \footnotesize
% % 1^{\tt st}\texttt{-step fusion}:
%     \mathtt{\mathbf{s}}_{\mathbb{P}} = \frac{\sum\nolimits_{{\tt p} \in \mathbb{P}} \mathtt{\mathbf{p}}}{|\mathbb{P}|}, \quad
%     \mathtt{\mathbf{s}}_{\mathbb{F}} = \frac{\sum\nolimits_{{\tt f} \in \mathbb{F}} \mathtt{\mathbf{f}}}{|\mathbb{F}|},
% \end{equation}
\begin{equation}
\footnotesize
% 1^{\tt st}\texttt{-step fusion}:
    \mathtt{\mathbf{s}}_{\mathbb{P}} = {\sum\nolimits_{{\tt p} \in \mathbb{P}} \mathtt{\mathbf{p}}}\big/{|\mathbb{P}|}, \quad
    \mathtt{\mathbf{s}}_{\mathbb{F}} = {\sum\nolimits_{{\tt f} \in \mathbb{F}} \mathtt{\mathbf{f}}}\big/{|\mathbb{F}|},
    \vspace*{-1em}
\end{equation}

% \begin{equation}
%     (\beta_{\phi_{1}}, ... ,\beta_{\phi_{P}}) = att_{sem}(Z_{\phi_{1}}, ... ,Z_{\phi_{P}})
% \end{equation}

\begin{equation}
\small
    % \texttt{procedure set:} 
    \mathbb{P} =\{{\tt m}|r({\tt m},{\tt n}) =\text{Next}\},
\end{equation}
\begin{equation}
\small
    % \texttt{factual set:} 
    \mathbb{F} =\{{\tt m}|r({\tt m},{\tt n}) \neq \text{Next} \land |r({\tt m},{\tt n}) |\neq 0\},
    \vspace*{-0.6em}
\end{equation}
where ${\tt n}$ is the current node, $\tt m$ denote another node in the graph, $\mathtt{\mathbf{n}}$ is the embedding of ${\tt n}$, $\mathbb{P}/\mathbb{F}$, $\mathtt{\mathbf{s}}_{\mathbb{P}}/\mathtt{\mathbf{s}}_{\mathbb{F}}$, $\alpha_{\mathbb{P}}/\alpha_{\mathbb{F}}$, and $\sum \mathtt{\mathbf{s}}_{\mathbb{P}}/\sum \mathtt{\mathbf{s}}_{\mathbb{F}}$ are the procedural/factual neighboring node set, procedural/factual set embedding, attention weight for procedural/factual information, and the sum of all procedural/factual set embeddings in the graph, respectively.  
% where $Z$ denotes weighted fused node embedding from set embeddings $Z_{\mathbb{P}}$ and $Z_{\mathbb{F}}$, which are obtained through fusing embeddings in the same neighboring nodes set, and the weights of fusion are calculated through attention mechanism over all set embeddings. 
In this way, each node is jointly embedded from both procedural and factual information associated with it.\par

Further, the user question is encoded as a vector using pre-trained language models. To bridge the semantic gap between a node (${\tt n}$) and the question~(${\tt q}$), we learn a matrix ${\tt W}$ to project the question embedding into the graph spaces. The alignment process can be formulated as $\mathop{\arg\max}_{{\tt n}} {\tt softmax} ( \mathtt{\mathbf{n}} \cdot \mathtt{\mathbf{W}} \cdot \mathtt{\mathbf{q}})$.
% \begin{equation}
% \small
%     \mathop{\arg\max}\limits_{n} {\tt softmax} ( \mathtt{\mathbf{n}} \cdot \mathtt{\mathbf{W}} \cdot \mathtt{\mathbf{q}}).
% \end{equation}
In other words, the question clue node is the most similar one to the user question. Notably, although we only align the user question to a single node in the graph, we believe this node is enough to cover all information in the user question. Because the embedding of a node has integrated adjacent arguments information and contains the core concept of the user question which is necessary for further inference. For instance,  the action node ``sign'' has integrated the information of its argument ``policy'' and can represent the complete action ``sign the policy'', which is sufficient for subsequent clue inference.

\subsection{Inference}
\label{sec:infor}
We model the inference of a possible response as the link prediction from the question clue node~($\tt n_q$) to its neighboring nodes. First, we recompose the question clue node as a question clue. For example, the ``align'' node is combined with its arguments and forms the question clue ``\textcolor{DarkOrchid}{Finally, sign the policy}''~($\tt Q_c$). This question clue is encoded as a vector using the same language model as the user question. As such, the required link information~($r$) can be estimated from $\mathtt{\mathbf{Q_c}}-\mathtt{\mathbf{n_q}}$. We then project the link information into the graph space for inference, denoted as $\mathtt{\mathbf{r}}=\mathtt{\mathbf{W}}\cdot(\mathtt{\mathbf{Q_c}}-\mathtt{\mathbf{n_q}})$. We further add a self-loop edge on $\tt n_q$ for the case, where the question clue node is also the response clue node.\par
We then form a triple with the question clue node as the head node~($\tt h$), a candidate node as the tail node~($\tt t$), and the link information as their relation~($\tt r$). The plausibility score of triple $<\tt h, r, t>$ is estimated through the adaptive convolution strategy~\cite{jiang2019adaptive}. It splits the link information~$\mathtt{\mathbf{r}}$ into procedural and factual matrixes, and uses them as filters on the head and tail nodes to produce convolutional features. The plausibility score is computed from the convolutional features. 
% which is designed to first produce multi-relational feature of head node and relation through convolution across the procedural and factual information by two filters, and then calculate the matching degree between the multi-relational feature and the tail node to estimate the plausibility score of triple $<\tt h, r, t>$.  
% which is defined as 
% \begin{equation}
% \label{equ:convr}
% \small
%     \psi(\mathtt{\mathbf{h, r, t}}) = \sigma ( \mathcal{f}(\mathtt{\mathbf{h}}, \mathtt{\mathbf{r}})^{\top} \cdot \mathtt{\mathbf{t}} ),
% \end{equation}
% where $\mathcal{f}$ denotes cross-convolution operation and $\sigma$ denotes the sigmoid function. 

Towards better coverage of possible responses, we conduct joint inference on both procedural and factual paths with the beam search strategy. It starts from the question clue node~(``sign''), walks on the procedural path~(``sign'' $\rightarrow$ ``see'' $\rightarrow$ ``agree to'') to find a response clue node and walks on factual paths~(``sign'' $\rightarrow$ ``User'' $\rightarrow$ ``browser'' and ``sign'' $\rightarrow$ ``User'' $\rightarrow$ ``real name authentication user'') to find the other two response clue nodes~(cf., Figure~\ref{fig:framework}, ).
% Starting from the question clue node, the model walks along both types of paths to infer possible response clue nodes. 
On either type of path, the inference ends when we meet nodes whose plausibility scores are above a pre-defined threshold $\delta$ or the maximum inference depth is reached. \par

% The entire inference process is summarized in Algorithm~\ref{algorithm:infer}. \textcolor{red}{more details} 
% Starting from the question clue node $n_{q}$, the model conducts inference across both procedural and factual paths to find responses to the questions. 

% Specifically, when the question is aligned to the action node, the model only walks along the procedural path because, during the encoding process, each action node has fused with other factual information from the nodes around it. 
% and inferences based on different paths are distinguished by different neighbor nodes selection functions, i.e., find neighbors linking with procedural or factual paths. 
Similar to the recomposition of question clues, the nodes on the path to the response clue node are recomposed as transitional clues, and the response clue node is recomposed as a response clue for highlighting. 
The paths from the
question clue node to the response clue node form the clue chains from the question clue to the response clue, which serve as explicit explanations in our \sysname assistant. 

% the nodes in clue chain from the user question to a response can be recomposed as clues in the same way. 

\begin{figure}[t]
    \centering
    \includegraphics[width=8cm]{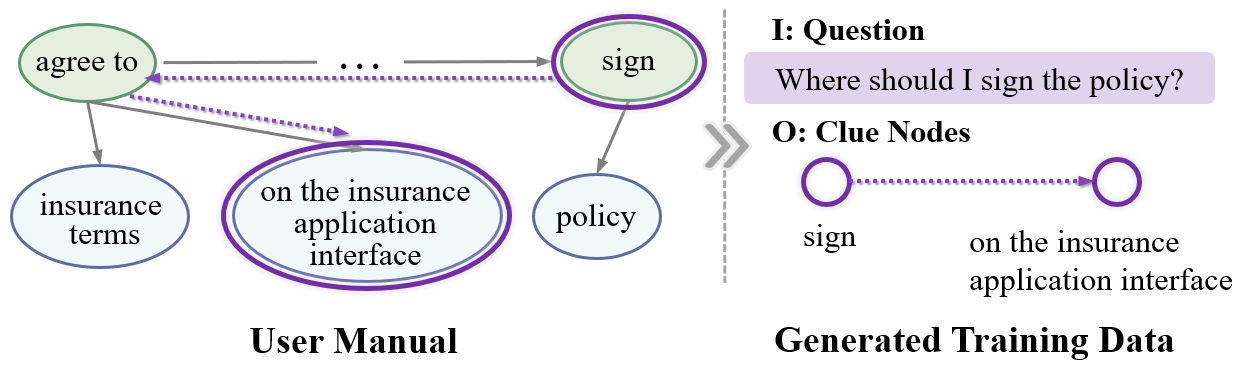}
    \vspace{-0.3cm}
    \caption{The construction of a sample for self-supervised training.}
    \label{fig:data_generation}
\end{figure}

\subsection{Self-supervised Training}
\label{sec:sst}
\paragraph{Data Construction}
To overcome the shortage of dedicated annotated data, we adopt the self-supervised strategy to enhance the reasoning ability of the model only with the inherent information of user manuals. Specifically, we first mask an argument node of an action or entity as the response clue node, then take a node that is not directly connected to it as the question clue node, and generate a natural language question that aligns with the question clue node and can be answered by the masked argument. As we simultaneously generate procedural and factual questions, this enhances the model's ability to answer complex questions that require joint reasoning on both procedures and facts binding them. Figure~\ref{fig:data_generation} presents a sample constructed in this way. Specifically, we collect $10,000$ user manuals from a Chinese online e-commerce platform and represent them as heterogeneous graphs. In total, we augment $43,348$ question-answer pairs, each of which is attached with a supporting user manual, for model training. 

% \begin{figure*}[!t]
%     \centering
%     \includegraphics[width=\textwidth]{figures/graph.png} 
%     \caption{An detailed example for the constructed graph of user manual.}
%     \label{fig:graph}
%      \vspace*{-0cm}
% \end{figure*}

\paragraph{Objective}
The objective of the backbone model is to minimize the sum of the binary cross entropy between the predicted clue nodes and the ground truth ones. Note that, during training, we only take the node with the highest plausibility score as the response clue node. During testing, there may be more than one response clue node. 

\section{Experiments}
An ideal reading assistant is expected to have two advantages. First, it should be efficient to highlight accurate responses to user questions. Thus, we design a response clue reasoning experiment to evaluate the ability of the backbone model of \sysname to infer accurate responses. Second, it should save CSRs' time at low risk by offering clue chains to guide the CSRs to arrive at the responses. We further evaluate \sysname in the real-world scenario by deploying it to train newly-hired CSRs, namely, the performance test under online deployment.

% \subsection{Settings}
\label{exp:settings}
% \paragraph{Training Corpus} To construct the self-supervised data for training, we collect $10,000$ user manuals from a Chinese online e-commerce platform and represent them as heterogeneous graphs. Following the strategy described in Section~\ref{sec:sst}, we get a total of $43,348$ valid samples. Each sample contains a question, the response to answer the question, and a referring user manual. The collected samples are used for self-supervised training of our backbone model. 

% \subsection{Implementation Details}
User manuals are mapped into graphs following \citet{liang2023knowing} with modifications defined in Sec.~\ref{graphs detail}.
We use the pre-trained PERT model\footnote{We use the implementation released by \url{https://github.com/ymcui/PERT}.}~\cite{cui2022pert} to embed user questions and initialize nodes embeddings. The threshold $\delta$ is set to $0.5$. The backbone model of \sysname is trained on two Nvidia P100 GPUs with the Adam optimizer. The learning rate, beam size, and maximum inference depth are set to $1e-5$, $4$, and $3$, respectively. 

\subsection{Response Clue Reasoning}
\paragraph{Task Description}
We construct a real-world testing set from a Chinese online e-commerce platform to test the backbone model's ability to infer accurate responses. It consists of $5,000$ manuals from $35$ fields, including finance, healthcare, insurance, electronic products, logistics, etc. 
% To test the model's generalization ability for manuals in different fields, we first selected $5,000$ manuals from the platform of up to $35$ fields, including finance, healthcare, insurance, electronic products, logistics, etc., containing all the fields covered by the platform. 
Each user manual is attached with its most frequently asked user question and the response to the question extracted from past conversations between CSRs and users. We then hire five senior CSRs to check the reliability of the extracted responses and correct minor errors~(e.g., typos). Finally, we obtain a testing set with $4,393$ reliable question-response pairs along with supporting user manuals. The response clue reasoning experiment is based on this testing set --- given a user question and its supporting user manual, the model needs to infer the response to the question from the manual. 
% For a fair comparison, our proposed model and baseline models only output one response with the highest prediction probability.

% \usepackage{multirow}
\begin{table*}[t]
\caption{Results of the response clue reasoning experiment. Best performances are highlighted \textbf{in bold}.}
\centering
\renewcommand\arraystretch{1.1}
% \resizebox{\linewidth}{!}{
\scalebox{0.8}{
\begin{tabular}{|l|c|c|c|c|c|c|}
\hline
\renewcommand\arraystretch{1.1}
% \textbf
{{Method}}                  & BLEU-1 & BLEU-2 & BLEU-3 & BLEU-4 & Rouge-L                         & BERTScore                       \\ \hline
% \textbf
{PERT~\cite{cui2022pert}}                    & 0.298 & 0.207 & 0.162 & 0.135 & 0.307                           & 0.251                           \\ \hline%\midrule[0.5pt]\specialrule{0em}{0pt}{2.2pt}
% \textbf
{Fine-tuned PERT}         & 0.433 & 0.345 & 0.292 & 0.254 & 0.438                           & 0.396                           \\ \hline%\midrule[0.5pt]\specialrule{0em}{0pt}{2.2pt}
% \textbf
{TARA~\cite{liang2023knowing}}                    & 0.397 & 0.301 & 0.239 & 0.195 & 0.361                           & 0.327                           \\ \hline%\midrule[0.5pt]\specialrule{0em}{0pt}{2.2pt}
% \textbf
{DRGN~\cite{zheng2022dynamic}}                    & 0.405 & 0.312 & 0.247 & 0.206 & 0.385                           & 0.356                           \\ \hline%\midrule[0.5pt]\specialrule{0em}{0pt}{2.2pt}
% \textbf
{ChatGPT~\cite{ouyang2022training}}                 & 0.418 & 0.322 & 0.266 & 0.229 & 0.418                           & 0.400                           \\ \hline%\midrule[0.5pt]\specialrule{0em}{0pt}{2.2pt}
\textbf{\sysname}                & \textbf{0.544} & \textbf{0.450} & \textbf{0.387} & \textbf{0.339} & \textbf{0.497} & \textbf{0.484} \\ 
\hline%\bottomrule[1.5pt]
\end{tabular}
}
\label{tab:qa}
% \vspace{-0.5cm}
\end{table*}

\paragraph{Baselines} 
We compare the proposed backbone model with five state-of-the-art baselines\footnote{For a fair comparison, all models are evaluated on the response with the highest prediction probability.}: 
\begin{itemize}
    \item \textbf{PERT}~\cite{cui2022pert}, which is a pre-trained machine comprehension model. It takes the concatenation of a user question and its supporting user manual as input and predicts the starting and ending locations of a possible response. 
    
    \item \textbf{Fine-tuned PERT}, which is also the PERT model but fine-tuned on our corpus~(cf., Sec.~\ref{exp:settings}). 
    % We input the concatenation of a pair of question and corresponding user manual to the model, and promote the model to predict the start and end points of the response in the manual. 
    
    \item \textbf{TARA}~\cite{liang2023knowing}, which, similar to our work, also represents user manuals as heterogeneous graphs but uses several manually defined rules to find responses.
    % defines several heuristic rules to enable response inference over the heterogenous graphs of user manuals. We construct the graphs through the approach described in~\ref{graphs detail}. Then we conduct the response inference in user manuals following the approach in the original paper. 
    
    \item \textbf{DRGN}~\cite{zheng2022dynamic}, which is a question-answering model for factoid knowledge graphs and fine-tuned on our training corpus. We form the factoid knowledge graph of a user manual by treating subjects and objects as entities and predicates as relations between them. 
    % The user manual is parsed by the N-LTP parsing tool~\cite{che2020n} 
    % A triple in the graph is
    % The knowledge graph is established by extracting entities and the relationships between them to form various triples and fusing entities with the same reference. We adopt the N-LTP parsing tool~\cite{che2020n} to extract triples from the user manuals. 
    Given a user question, the DRGN model aligns it to an entity node in the graph and walks on the graph to infer a possible response.
    % the node where the response to the question is located through reasoning on the constructed knowledge graph. 
    
    \item \textbf{ChatGPT}~\cite{ouyang2022training}, which is instructed to infer responses from the user manual via prompts. We adopt the Tree of Thought (ToT)~\cite{yao2023tree} strategy to infer the response step by step. We first prompt ChatGPT to find actions/entities aligned with the user question, then prompt it to infer the response clues step by step for each aligned action or entity, until reach a possible response to the question.  
\end{itemize}

\paragraph{Metrics}
Following \citet{cui2022pert} and \citet{zheng2022dynamic}, we adopt BLEU~\cite{papineni2002bleu}, ROUGE-L~\cite{lin2004rouge} and BERTScore~\cite{zhang2019bertscore} as metrics.
Both the BLEU scores and Rouge-L score measure the lexical similarity between the predicted response with the gold one. The difference is that the BLEU scores focus more on the precision of predicted responses, but the Rouge-L score focuses more on the recall rate of the predicted responses. Besides, the BERTScore score focuses more on the semantic similarity. 

% of the predicted responses. 
% All of the above metrics are calculated based on the model predicted responses and corresponding annotated gold responses. 

\paragraph{Results}
The experimental results are shown in Table~\ref{tab:qa}. Compared with the baselines, our backbone model gets the best scores on all metrics. The results demonstrate that our backbone model is able to integrately  understand both procedural and factual information in user manuals and infer accurate responses to the user questions. Moreover, the joint inference on both procedural and factual paths in the graphs with the beam search strategy further improves the robustness of the backbone model for inferring the desired responses. Additionally, we have the following observations:

1) Both TARA and our backbone model represent user manuals as heterogeneous graphs. However, the performances of TARA are much worse than ours. This is because compared with the hand-craft rules of TARA, our backbone model is more powerful in making unified inferences of actions and entities and thus has better coverage of various user questions in real-world scenarios. 
What's more, our backbone model is trained on the constructed training corpus with various generated questions through the self-supervised strategy. This greatly improves the model's generalization ability to cope with complex user questions.

2) DRGN, TARA, and our backbone model all conduct inference on structured graphs of user manuals. DRGN outperforms TARA but still performs much worse than ours. This is because DRGN obtains a more comprehensive understanding of user manuals through self-supervised learning, and TARA is limited by manually designed inference rules. However, DRGN still fails to achieve good performance as it only focuses on the factual information of user manuals. Our backbone model, by contrast, emphasizes both procedural and factual information in manuals and is capable of supporting joint inference to find proper responses. 

3) PERT performs the worst among all baselines due to its inability to conduct joint inference on structured data and its lack of knowledge related to the user manuals during the pre-training process. 
Although both PERT and ChatGPT have not been fine-tuned on the training corpus we constructed, ChatGPT uses a larger volume of pre-trained corpus covering a wide range of fields and thus achieves better performance than PERT. 

4) Fine-tuned PERT gains significant performance improvement through learning the inherent information of user manuals, which indicates that our proposed self-supervised learning strategy can greatly improve the model's ability to infer the responses to user questions. 
However, the fine-tuned PERT model still performs worse than ours due to a lack of joint inference ability. Our backbone model benefits from joint inference on both procedural and factual paths and thus can provide more accurate responses. 

5) ChatGPT is good at document comprehension, but its performance is still worse than ours. This is because ChatGPT does not have the explicit reasoning ability, and due to its inherent hallucination issues, it often outputs factually incorrect responses. This poses a significant risk of misleading CSRs in practical scenarios. An ideal reading assistant is expected to save CSRs' time at low risk by offering clue chains to guide the CSRs to arrive at the responses. 
% Even with the Tree of Thought strategy, ChatGPT can still output incorrect reasoning chains and provide factually incorrect answers, which 
Our backbone model overcomes this challenge by explicitly performing joint inference on heterogeneous graphs. Therefore, the predicted responses of our backbone model are interpretable with inference chains and thus are more practical in real-world applications. 

6) Moreover, ChatGPT performs even worse than fine-tuned PERT, this is because the fine-tuned PERT model learns semantic associations between various user questions and corresponding responses during the fine-tuning process, thus can handle more complex user questions. The responses generated by ChatGPT have better semantic coherence, thus slightly outperforming fine-tuned PERT under the BERTScore metric. 
\begin{figure}
    \centering
    \includegraphics[width=6.5cm]{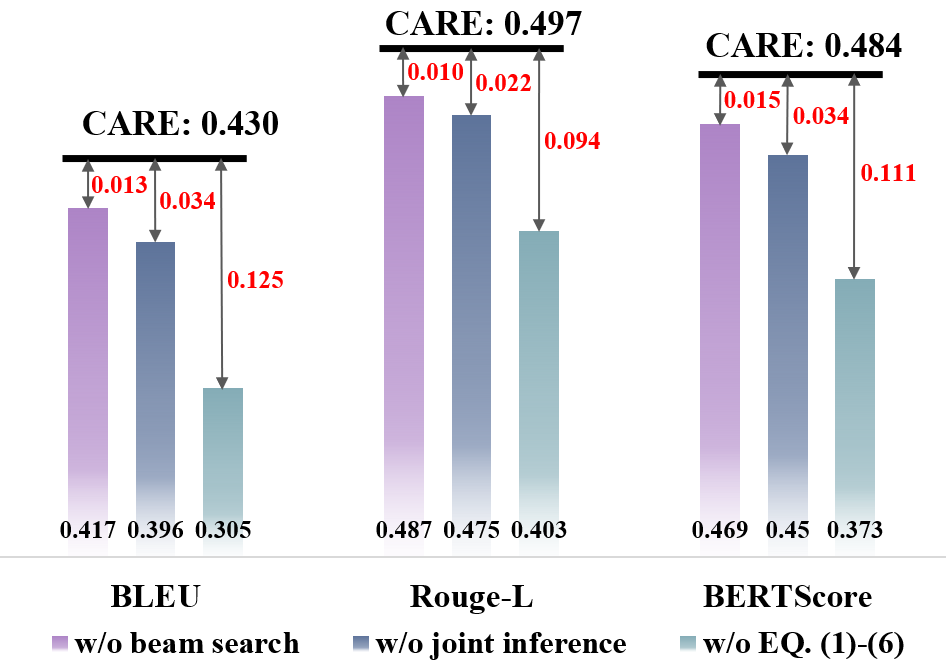}
    \caption{Results of the ablation study}
    \label{fig:ablation}
    % \vspace{-1em}
\end{figure}

\paragraph{Ablation Study}
The results of ablation experiments are shown in Figure~\ref{fig:ablation}, demonstrating the effectiveness of proposed strategies in our backbone model. Note that, the BLEU score here is the average score of BLEU-1, BLEU-2, BLEU-3, and BLEU-4 scores. There is a decline in performance of ``w/o beam search'' compared with \sysname. This is because, without the beam search strategy, the accuracy of the response clue node prediction will be heavily affected by the errors accumulated from the previously predicted clue nodes. It can be seen that the performance of ``w/o joint inference'' drops because, after replacing the joint inference~(cf., Sec. \ref{sec:infor}) with direct link prediction to all nodes in the graph, the model loses the ability to reason between procedural and factual information by walking on both procedural and factual paths. 
We also compare the backbone model of \sysname with a model ``w/o EQ.~(1)-(6)''. This model encodes the heterogeneous graph in a homogeneous way, where the neighboring node set of a node containing all nodes related to it. There are significant performance decreases of ``w/o EQ.~(1)-(6)'' compared with \sysname. This is because, without the unified representation defined in EQ.~(1)-(6), the model loses the ability to integratedly comprehend the user manuals from both procedural and factual knowledge. 

% Without the beam search strategy, the accuracy of the response clue node prediction is affected by the previous clue nodes prediction error, leading to a performance decline. 
% After replacing the joint inference with direct link prediction to all nodes in the graph, the performance decreases because the model loses the ability to reason between procedural and factual information by walking on both procedural and factual paths. 
% When ignoring the different types of nodes and edges in the graph~(i.e. without heterogeneous reasoning strategy), the performance of the model drops sharply because, if not synthetically considering both the procedural and factual information in the graph, the model can not reach a comprehensive understanding to the user manuals. 

% It can be seen from the performance decline of the first column under each metric in Figure~\ref{fig:ablation} that, without the beam search strategy, the accuracy of the response clue node prediction is affected by the previous clue nodes prediction error.
% From the performance decline of the second column under each metric, we see that the model loses the ability to reason between procedural and factual information by walking on both procedural and factual paths after replacing the joint inference with direct link prediction to all nodes in the graph.

\subsection{Online Experiment}
\paragraph{Task Description} We further conduct a live test in the real-world scenario, where newly hired CSRs are trained with the help of reading assistants. Under the between-subjects setting~\cite{charness2012experimental}, we invite three groups of newly hired CSRs, each consisting of two individuals. All of them are asked to answer user questions based on corresponding manuals, with the first group using \sysname, the second group using the assistant only highlighting response clues, and the third group using the assistant presenting raw user manuals. We also compare these assistants with the backbone model of \sysname. We select $516$ high-frequency access samples from the CSR training bank for testing. 

\paragraph{Metrics} Besides the BLEU, Rouge-L, and BERTScore scores, we also care about whether it can save time with the help of \sysname. Hence, we also report the average time spent for a CSR to answer all user questions. To further examine the reliability of human results, we calculate the ICC~(Intraclass Correlation Coefficient)~\cite{shrout1979intraclass} score for each group. The ICC score ranges from 0 to 1, with a high ICC close to 1 indicating high similarity between values from the same group and a low ICC close to zero meaning that values from the same group are not similar. According to \citet{koo2016guideline}, the ICC score, bigger than $0.75$, interprets that the values are in good reliability.

\begin{figure}[t]
    \centering
    \includegraphics[width=7cm]{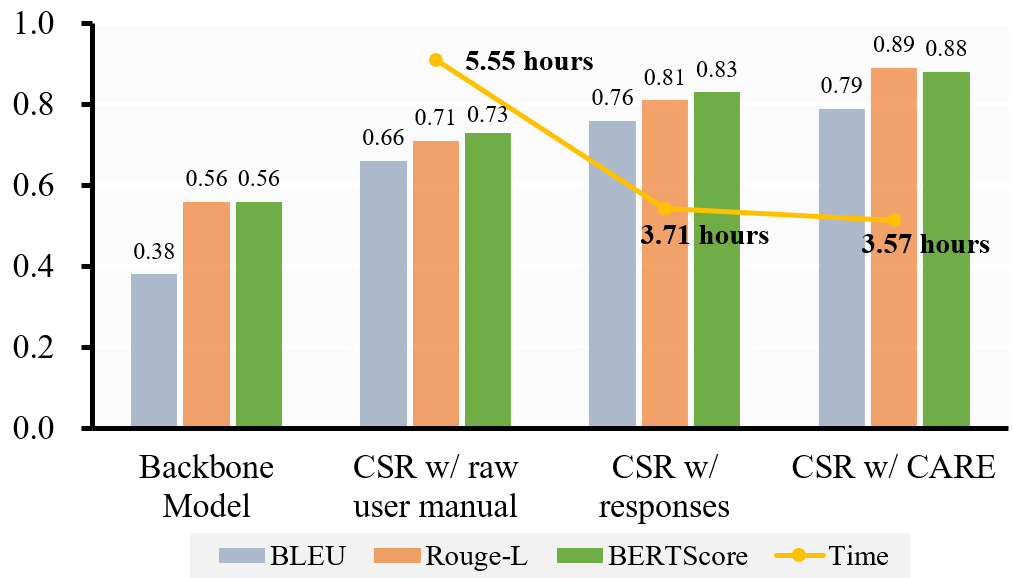}
    \vspace*{-0.1cm}
    \caption{Results with different reading assistants.}
    \label{fig:online}    
\end{figure}

\paragraph{Results}
% The reliability of the results is shown by the responses provided by various CSRs in these groups having ICC score~\cite{shrout1979intraclass} 0.8, 0.76 and 0.75 

The results are shown in Figure~\ref{fig:online}.
With the help of \sysname, the accuracy of the responses provided by CSRs has improved significantly compared to responses given by other CSRs receiving no assistance, and the assistant saves $35.7\%$ the time spent answering user questions for CSRs while keeping a $0.755$ ICC score with the gold responses and a $0.801$ ICC score with the new CSRs w/ raw user manual. 
% while keeping a high ICC score
This is because our assistant can guide the CSRs to the proper responses and explain them by showing clue chains, which reduces the reading burden of CSRs and improves the efficiency of answering user questions. 
CSRs receiving no assistance are easily biased by the content at the beginning of the manual due to the heavy reading burden, thus although gaining high consistency, the provided responses miss important response clues in the rest of the user manual. 
We can also see that the complete display of inferring clue chains can better improve the efficiency and service quality of CSRs, as providing explanations can better guide CSRs to understand the predicted responses and thus they can quickly decide to copy the predicted ones or find the responses themselves. Additionally, relying solely on the responses generated by the backbone model, though answering user questions automatically, gets the worst performances. There is no chance for a CSR to take remedial action for responses wrongly predicted by the model and thus it is too risky to be applied in online custom services. \par

\section{Conclusion}
In this paper, we propose a clue-guided assistant \sysname to help CSRs reduce the burdens of reading user manuals by providing accurate responses and explicitly showing explainable paths about how to arrive at these responses. Specifically, we design a backbone model that aligns user questions to the nodes in the heterogeneous graphs constructed from the manual and infers response clues through joint reasoning along with both procedural and factual paths. In addition, we alleviate the shortage of dedicated annotated data by conducting self-supervised learning on the inherent information of user manuals. 
Experimental results show that our proposed backbone model can provide accurate response clues for user questions, and the \sysname assistant can greatly save the time for CSRs to answer questions, thereby improving the efficiency and quality of online customer service. 
We have deployed our assistant to the novice CSRs training scenario and are preparing to deploy it to the online customer service applications in the future. 
% so that more CSRs can benefit from

% \clearpage
\section{Ethical Considerations}
In this paper, we present a clue-guided reading assistant to help CSRs read user manuals and find proper responses to user questions more quickly. All of the user manuals, questions, and clues involved in our experiments are collected from an e-commerce platform and the collected data in our work does not contain any personal or sensitive information. Therefore, we believe that there are no ethical issues with our work.

\section{Limitations}
Manually annotating heterogeneous graphs is time-consuming and laborious, so we do not evaluate the adequacy of the constructed heterogeneous graphs. Instead, we evaluate the heterogeneous graphs' effectiveness through their assistance in helping to find accurate responses to user questions and provide explicit explanations of these responses in our experiments. Although the user manuals adopted in our experiments are collected from one online e-commerce platform, they cover various fields, including finance, healthcare, etc. We believe the evaluation results not only demonstrate the effectiveness of \sysname in the e-commerce CSR setting but show the potential of \sysname in other CSR settings where CSRs need to frequently read information-rich user manuals. The ideal online experiment is to deploy \sysname in real service scenarios to test its performance. However, this is not practical as no company is willing to take the danger of making wrong answers, as it may cause poor user experience and may even lose potential customers. As a remedy,  we deployed \sysname in a relative risk-free environment, where newly hired CSRs are trained with the help of reading assistants. 

\section*{Acknowledgement}
This work was supported in part by the National Natural Science Foundation of China (No. 62206191 and No. 62272330);
in part by the Natural Science Foundation of Sichuan (No. 2023NSFSC0473), 
and in part by the Fundamental Research Funds for the Central Universities (No. 2023SCU12089 and  No. YJ202219).

% \bibliography{custom}

\bibliography{anthology}

\clearpage
\begin{figure*}[t]
    \centering
    \includegraphics[width=\textwidth]{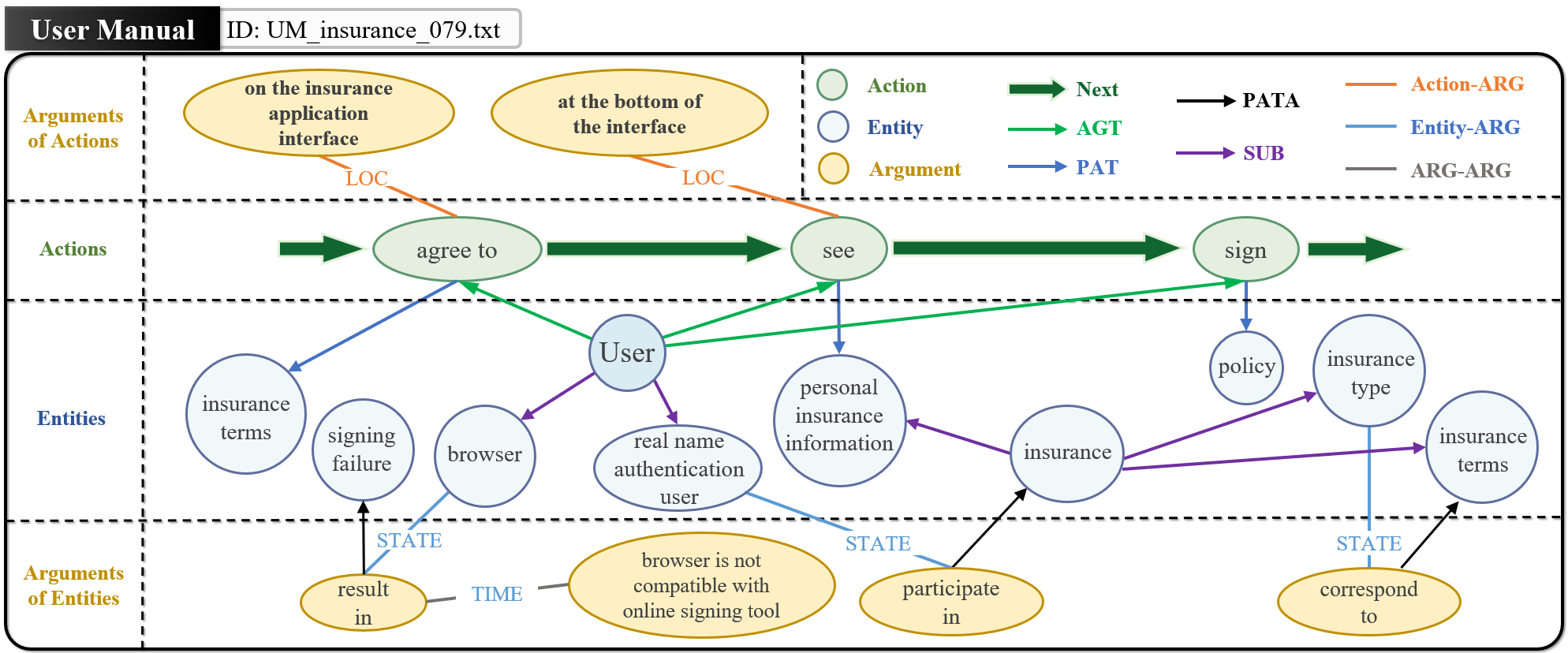} 
    \caption{An example of the constructed graph.}
    \label{fig:graph_appendix}
\end{figure*} 

\appendix

\section{Details of the Constructed Graphs}
\label{graphs detail}
Here we describe the constructed graphs of user manuals and explain how can the model infer the desired clues of user questions on the graph through joint reasoning on both procedural and factual information.

\subsection{Graph Structure of User Manuals}
We construct user manuals into graphs as shown in Figure~\ref{fig:graph_appendix}. 
The graph consists of three kinds of nodes (i.e. action nodes, entity nodes, and argument nodes) and eight kinds of relations (i.e. Next, AGT, PAT, PATA, SUB, Action-ARG, Entity-ARG, and ARG-ARG). 
Firstly, the procedures containing a list of action nodes are constructed by linking actions belonging to the same procedure with the ``Next'' relation. 
We use the Next Sentence Prediction(NSP) of the pre-trained model~\cite{shi2019next} to determine whether two actions belong to the same procedure and construct all the procedures. 
To establish connections between agents and actions, we employ the "AGT" relation, while the "PAT" relation is utilized to connect patients with their respective actions. And the arguments of each action are linked through the ``Action-ARG'' relation, including the time, location, and manner description of the corresponding action. 
Secondly, we construct entity nodes and link their corresponding arguments through the ``Entity-ARG'' relation. The sub-entity of a certain entity is connected to it through a ``SUB'' relation. We also link the state of the entity with corresponding entities affected by it through the ``PATA'' relation and link the arguments associated with other entities' arguments through the ``ARG-ARG'' relation. 
Finally, We adopt character level matching to confirm the same entities and arguments references and fuse the same reference nodes into one node to avoid duplicate reference nodes of the constructed graphs. 

To support the heterogeneous inference of our proposed model, we retain the original three node types (e.g. ) and classify edge types into procedural and factual links. The former only includes the Next relationship in the graph, while the remaining relationships belong to factual link
\begin{figure*}[t]
    \centering
    \includegraphics[width=0.7\textwidth]{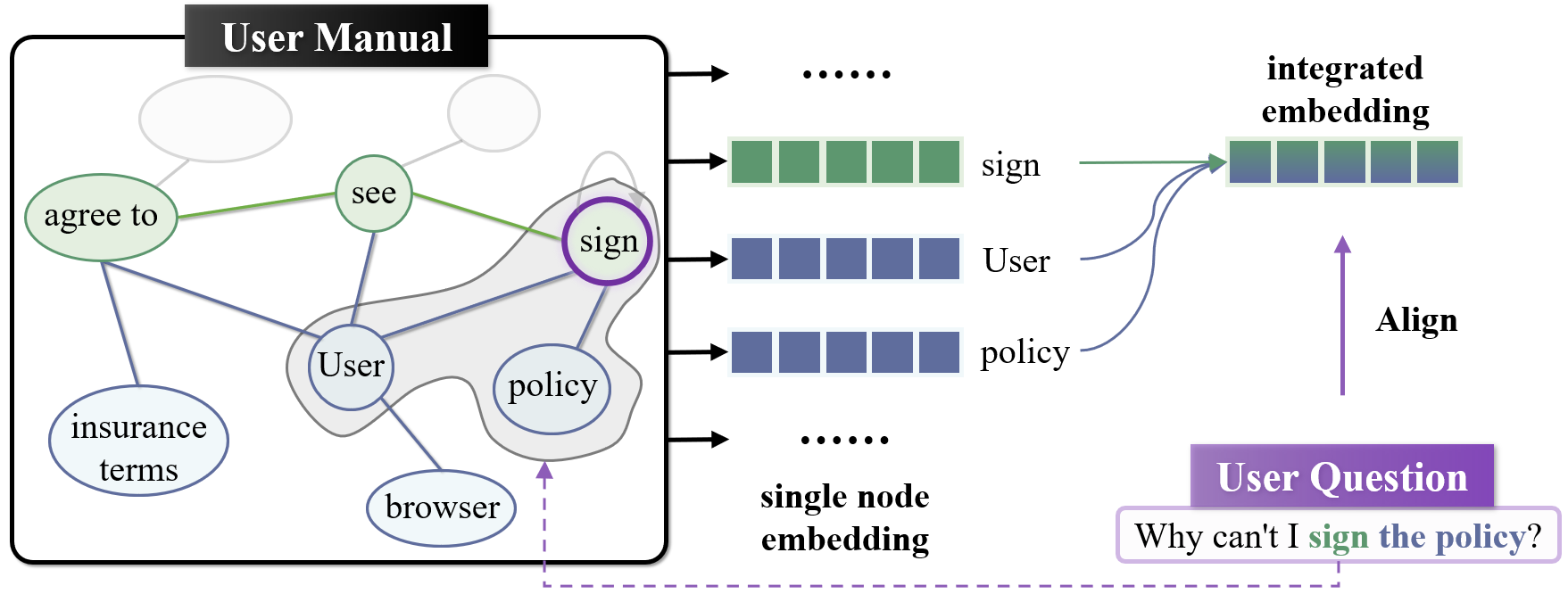} 
    \caption{An example of the integrated information of node ``sign''.}
    \label{fig:integrate}
\end{figure*} 
\subsection{Response Clues Inference}
We construct the graphs of user manuals and adopt heterogeneous desired clues inference of user questions.
With the constructed graph, our proposed model can perform joint reasoning on both procedural and factual information for sufficient and robust inference. 
As shown in Figure~\ref{fig:graph_appendix}, to answer the user question ``Why can't I sign the policy?'', the model first needs to identify the core action ``sign'' that the user question is targeting as the question aligned node for subsequent inference of the desired clues. 
It should be noted that the embedding of each node on the graph has integrated adjacent arguments information that is necessary for further inference. As shown in Figure~\ref{fig:integrate}, the action node ``sign'' has integrated the information of its argument ``User'' and ``policy'', and can represent the complete action ``User sign the policy'', which is sufficient for subsequent clue inference. 
Starting from the question-aligned node, the model conducts in-depth reasoning on the graph of the manual to find potential clues to the user question. Through joint reasoning on both procedural and factual information, the model first refers to the previous action ``agree to'' of the aligned node through procedural meta-path based inference and finds the potential clue for the failure of the action ``sign'', that is the user did not complete the prerequisite operation ``agreeing to the insurance terms on the insurance application interface'', resulting in the inability to ``sign the policy''. Then the model further infers to the restrictions ``Only real name authenticated users can participate in the insurance'' associated with ``user'' node's sub-entity ``real name authentication user'' and ``User browsers that are not compatible with online signing tools may result in signing failure'' associated with ``user'' node's sub-entity ``user browser'' through factual meta-path based inference, which may also result in the failure of the action ``sign''. 
Through the constructed graph, our proposed model can perform joint reasoning on both procedural and factual information to find potential clues and guide CSRs to reach the responses to user questions.

\section{Analysis of the Augmented Data}
\label{generated data analysis}

\begin{figure}[h]
    \centering
    \includegraphics[width=7.8cm]{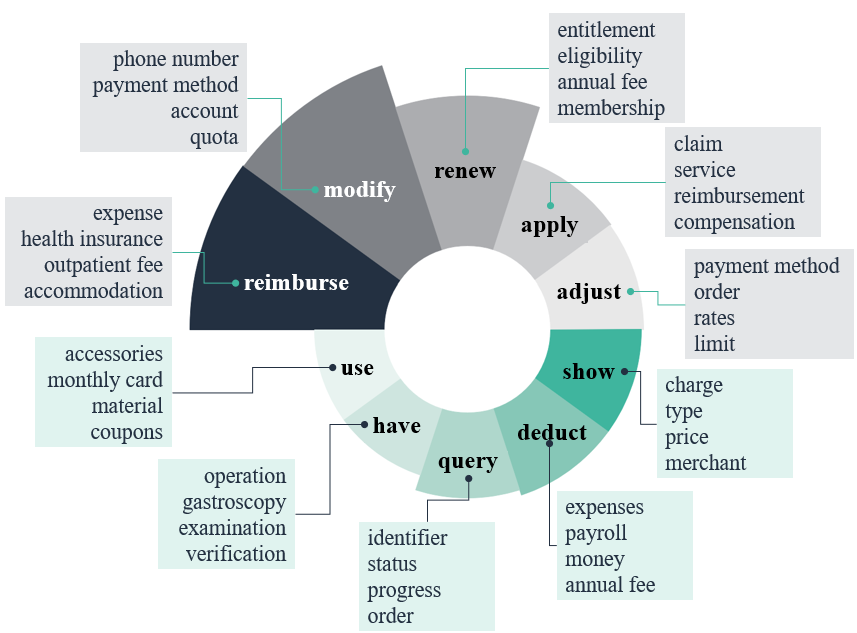}
    \vspace{-0.3cm}
    \caption{The top-$10$ actions and their top-$4$ patients of the questions in the augmented data.}
    \label{fig:question_analysis}
     % \vspace*{-0.5cm}
\end{figure}

% 43348  30670    4010 13.07 %  9.25%
% To improve the generalization ability of the model, we strive to ensure the diversity of generated questions during the process of training data generation. 
We demonstrate the diversity of augmented data by calculating the action and patient distribution of questions. After syntactic parsing\footnote{We use the syntactic parsing tool~\cite{che2020n}.}, we get $30,670$ questions holding ``action-patients'' structure. Specifically, the actions are verbs directly connected to the root in parsing trees and their patients are noun objects directly connected to the actions. 
% We parse the question into 
% We use the syntactic parsing tool~\cite{che2020n} to count the actions (verbs directly connected to the root in parsing trees) and corresponding patients (noun objects directly connected to the actions) in the generated questions. There are a total of $30,670$ questions with such a structure, while other factual style questions do not have such an ``action-patient'' structure (e.g. ``how much is the annual fee of membership benefits?''). 
We present the top-$10$ actions and their top-$4$ patients of the generated questions in Figure~\ref{fig:question_analysis}. It can be seen that the generated questions cover a wide range of actions, with a total of $4,010$ samples displayed, accounting for only $9.25\%$ of all generated questions and $13.07\%$ of questions with the ``action-patient'' structure. 
These generated procedural and factual questions enhance the model's reasoning ability of response clues for various questions raised by users in real-world scenarios and can improve the generalization ability of the model. 

\section{Computation Details of the Plausibility Score of Triple}
We model the inference of a possible response as the link prediction from the question clue node~($\tt n_q$) to its neighboring nodes. The question clue is encoded as a vector using the same language model as the user question and the required link information~($r$) can be estimated from $\mathtt{\mathbf{Q_c}}-\mathtt{\mathbf{n_q}}$. We project the link information into the graph space for inference, denoted as $\mathtt{\mathbf{r}}=\mathtt{\mathbf{W}}\cdot(\mathtt{\mathbf{Q_c}}-\mathtt{\mathbf{n_q}})$. Then we form a triple with the question clue node as the head node~($\tt h$), a candidate node as the tail node~($\tt t$), and the link information as their relation~($\tt r$). The plausibility score of triple $<\tt h, r, t>$ is estimated through the adaptive convolution strategy~\cite{jiang2019adaptive}. It splits the link information~$\mathtt{\mathbf{r}}$ into procedural and factual matrixes, and uses them as filters on the head and tail nodes to produce convolutional features. The plausibility score is computed from the convolutional features. 

% which is designed to first produce multi-relational feature of head node and relation through convolution across the procedural and factual information by two filters, and then calculate the matching degree between the multi-relational feature and the tail node to estimate the plausibility score of triple $<\tt h, r, t>$.  
% which is defined as 
% \begin{equation}
% \label{equ:convr}
% \small
%     \psi(\mathtt{\mathbf{h, r, t}}) = \sigma ( \mathcal{f}(\mathtt{\mathbf{h}}, \mathtt{\mathbf{r}})^{\top} \cdot \mathtt{\mathbf{t}} ),
% \end{equation}
% where $\mathcal{f}$ denotes cross-convolution operation and $\sigma$ denotes the sigmoid function. 

We first reshape question clue node feature $\mathbf{n_q} \in \mathbb{R}^{d_{n}}$ into 2D matrix $\mathbf{S} \in \mathbb{R}^{d_{n}^{h} \times d_{n}^{w}}$ and feed it into CNN layer. Here $d_{n} = d_{n}^{h} d_{n}^{w}$. 
To handle complex questions require multiple clues, we split relation feature $\mathbf{r} \in \mathbb{R}^{d_{r}}$ into same size pieces $\mathbf{r}^{(1)}, ..., \mathbf{r}^{(c)}$ (each $\mathbf{r}^{(l)} \in \mathbb{R}^{d_{r}/c}$) and reshape each $\mathbf{r}^{(l)}$ into 2D filter $\mathbf{R}^{(l)} \in \mathbb{R}^{h \times w}$. Here $c$ denotes the number of filters, $h$ denotes the height of each filter and $w$ denotes the width of each filter. 
After that, multiple relation-specific convolutions are conducted between matrix $\mathbf{S}$ and each filter $\mathbf{R}^{(l)}$: 

\vspace{-0.5cm}
\begin{equation}
    C^{l} = g(\mathbf{S} \ast \mathbf{R}^{(l)})
\end{equation}

\noindent
where $C^{l}$ denotes a feature map derived from filter $\mathbf{R}^{l}$, $g$ denotes the ReLU activation function and $\ast$ denotes convolution operation. 
To estimate the plausibility score of triple $<\tt h, r, t>$, we concatenate all feature maps $C^{1}, ...,C^{c}$ into single feature $\mathbf{c}$ and project it into $\mathbb{R}^{d_{n}}$ through fully-connected layer. The plausibility score of the triple can be calculated as follows: 

% \vspace{-0.5cm}
\begin{equation}
    \scalebox{.9}{$score(<\tt h, r, t>) = \mathtt{f}(\mathtt{g}(\mathbf{W}_{s} \cdot \mathbf{c} + \mathbf{b}_{s})^{\top} \cdot \mathbf{n_q})$}
\end{equation}

\noindent
where $\mathtt{f}$ denotes the sigmoid function, $\mathtt{g}$ denotes the ReLU activation function, $\mathbf{W}_{s}$ and $\mathbf{b}_{s}$ are fully-connected layer parameters.

\section{Details of Adopted Metrics}
The BLEU~\cite{papineni2002bleu} metric is used to evaluate the quality of generated text by measuring the n-gram overlap between the candidate and the reference texts, which is calculated as:

\begin{equation}
    \text{BLEU} = \text{BP} \times \exp\left(\sum_{n=1}^{N} w_n \log (p_n)\right)
\end{equation}

\noindent
where $\text{BP}$ is the brevity penalty, $N$ is the maximum n-gram size, $w_n$ is the weight for each n-gram size and $p_n$ is the precision for each n-gram size. 

The ROUGE-L~\cite{lin2004rouge} metric is used to evaluate the quality of generated text by measuring the longest common sub-sequence between the candidate and the reference texts, which is calculated as:

\begin{equation}
    R_{LCS} = \frac{\text{LCS}(r, c)}{\text{len}(r)}
\end{equation}

\begin{equation}
    P_{LCS} = \frac{\text{LCS}(r, c)}{\text{len}(c)}
\end{equation}

\begin{equation}
    \text{ROUGE-L} = \frac{(1 + \beta^{2}) R_{LCS} P_{LCS}}{R_{LCS} + \beta^{2} P_{LCS}}
\end{equation}

\noindent
where $\text{LCS}(r, c)$ is the length of the longest common sub-sequence between reference text $r$ and candidate text $c$, and $\text{len}(r)$ is the length of the reference text $r$, $\text{len}(c)$ is the length of the candidate text $c$, $R_{LCS}$ represents recall rate, $P_{LCS}$ represents precision rate, $\beta$ is used to adjust the attention to precision and recall rates. Due to $\beta$ being set to a large number, $\text{ROUGE-L}$ more considers recall rate. 

The BERTScore~\cite{zhang2019bertscore} metric is used to evaluate the quality of generated text by comparing it to reference text using contextualized embeddings generated by the BERT model. It computes the similarity between the contextualized embeddings of the candidate and reference text, and aggregates the scores to provide an overall BERTScore. 

The ICC~(Intraclass Correlation Coefficient)~\cite{shrout1979intraclass} score ranges from 0 to 1, with a high ICC close to 1 indicating high similarity between the evaluation values from the same group, which is calculated as:

\begin{equation}
    \text{ICC} = \frac{\text{MS}_{\text{between}} - \text{MS}_{\text{within}}}{\text{MS}_{\text{between}} + (k-1) \times \text{MS}_{\text{within}}}
\end{equation}

\noindent
where $\text{MS}_{\text{between}}$ is the mean square for between groups variability, $\text{MS}_{\text{within}}$ is the mean square for within groups variability and $k$ is the number of groups.

\section{Error Analysis}
As the focus of our work lies in the utilization of heterogeneous graphs, we will briefly discuss the error analysis of graph constructing of our method here. 
For the whole pipeline of our proposed approach, only the process of heterogeneous graphs constructing from the user manuals relies on the semantic parsing results of our chosen NLP tool LTP~\footnote{\url{https://ltp.ai/}}, and the parsing errors may have an impact on the performance of our proposed approach~(e.g., identify wrong argument node). 
However, after investigating the analysis of other parsing-enhanced approaches~\cite{sun2020sparqa, zhang2021fact, liu2022uni}, we conjecture these errors will not make a big difference in the model performances and corresponding conclusions. 
Furthermore, we believe that as fundamental NLP components advance, the influence of these errors on subsequent work will even be smaller.

\section{Discussion of the Used Artifacts}
All the codes, data and models used in this paper follow the settings of the original work, and we have listed the citations of all the artifacts we used in References section. All of the user manuals, questions, and clues involved in our experiments are collected from a Chinese online e-commerce platform.
All the collected data in our work has been manually desensitized and does not contain any personal or sensitive information, and all the data we use is for research purpose only.

\begin{figure*}[!th]
    \centering
    \subfigure[Co-reference Contents Highlighting]{\includegraphics[width=.45\textwidth]{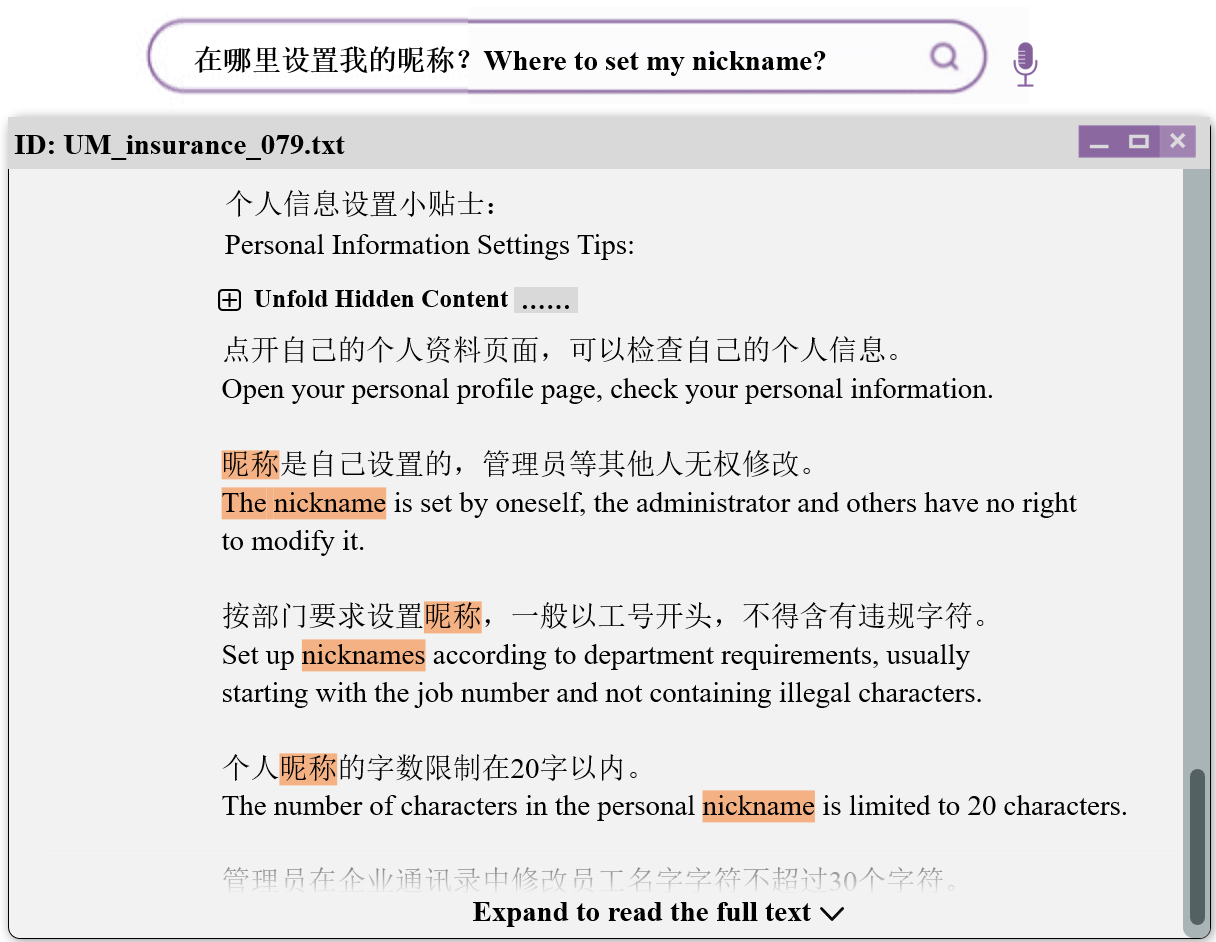}\label{fig:case1.4}}\hspace{1cm}
    \subfigure[Highlighting Specific Content]{\includegraphics[width=.45\textwidth]{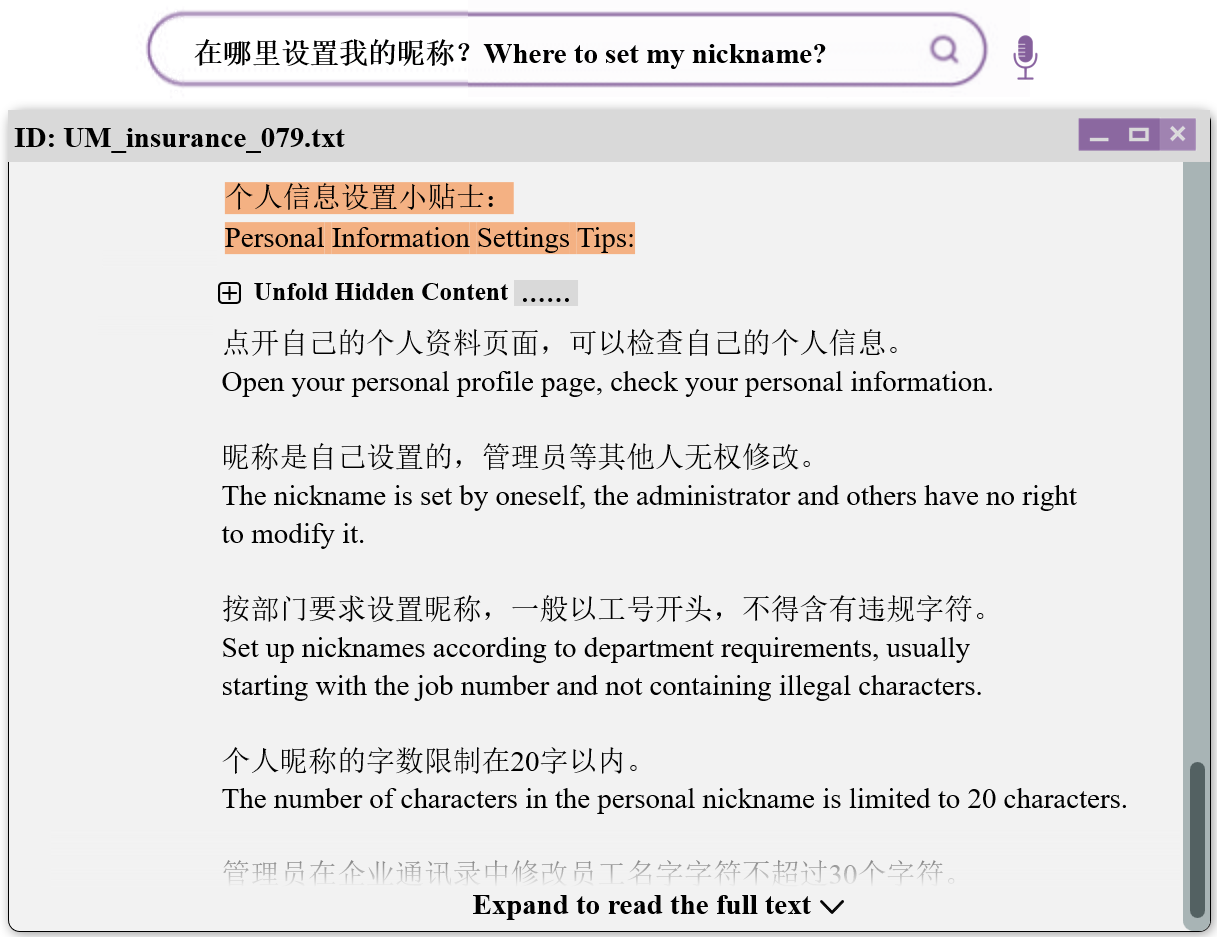}\label{fig:case1.5}}

    \vspace{1cm}
    \subfigure[Raw User Manual]{\includegraphics[width=.45\textwidth]{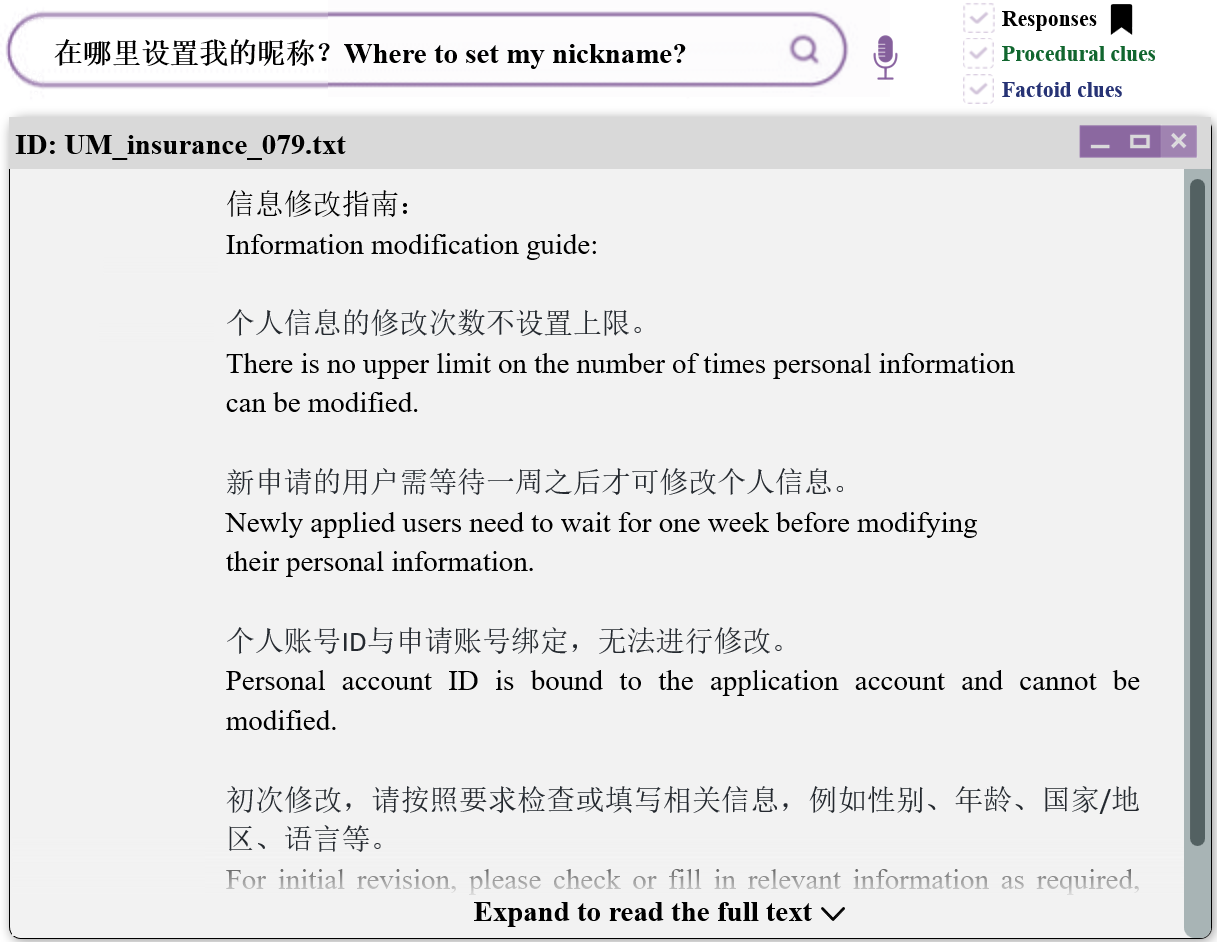}\label{fig:case1.1}}\hspace{1cm}
    \subfigure[Response Only]{\includegraphics[width=.45\textwidth]{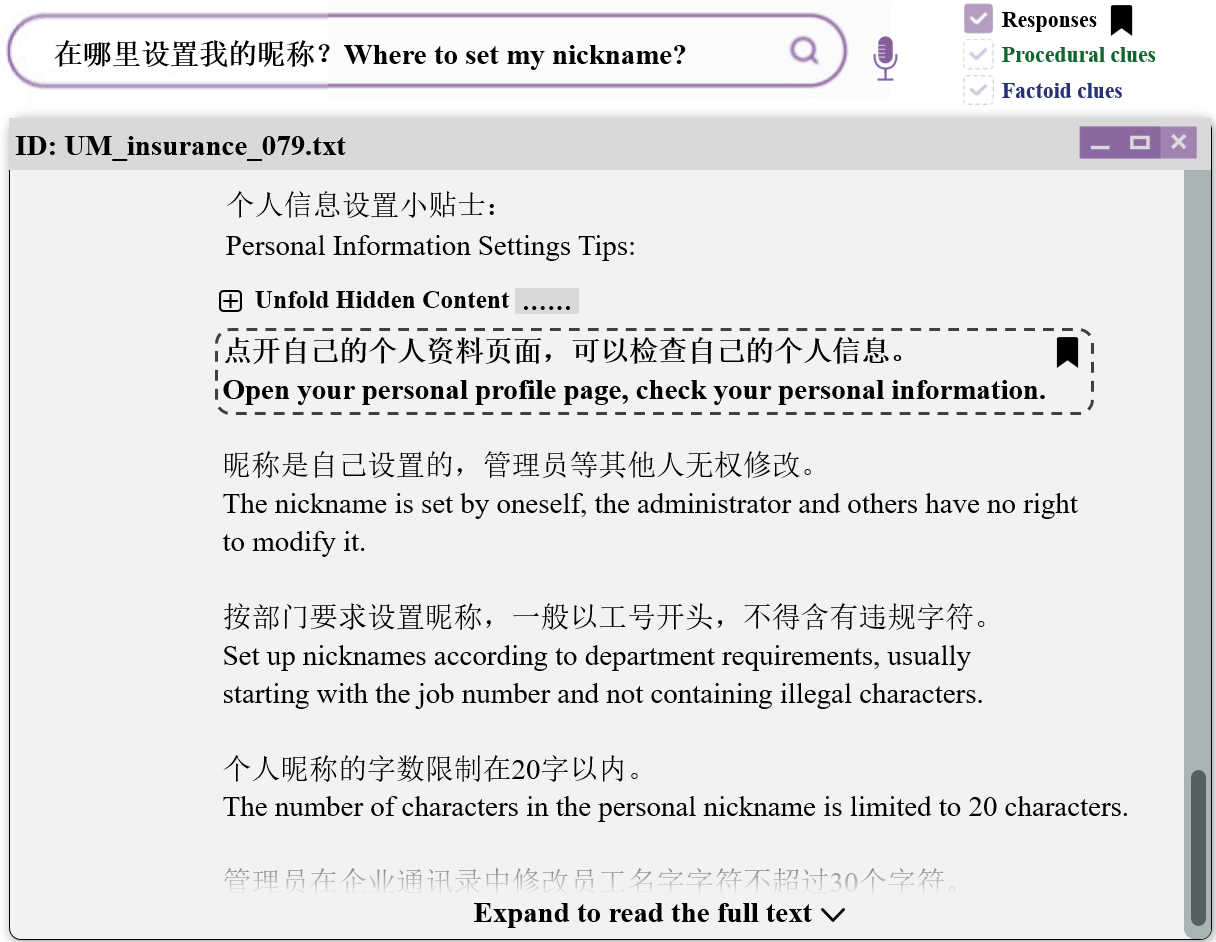}\label{fig:case1.2}}

    \vspace{1cm}
    \subfigure[Clue-guided Assistant]{\includegraphics[width=.45\textwidth]{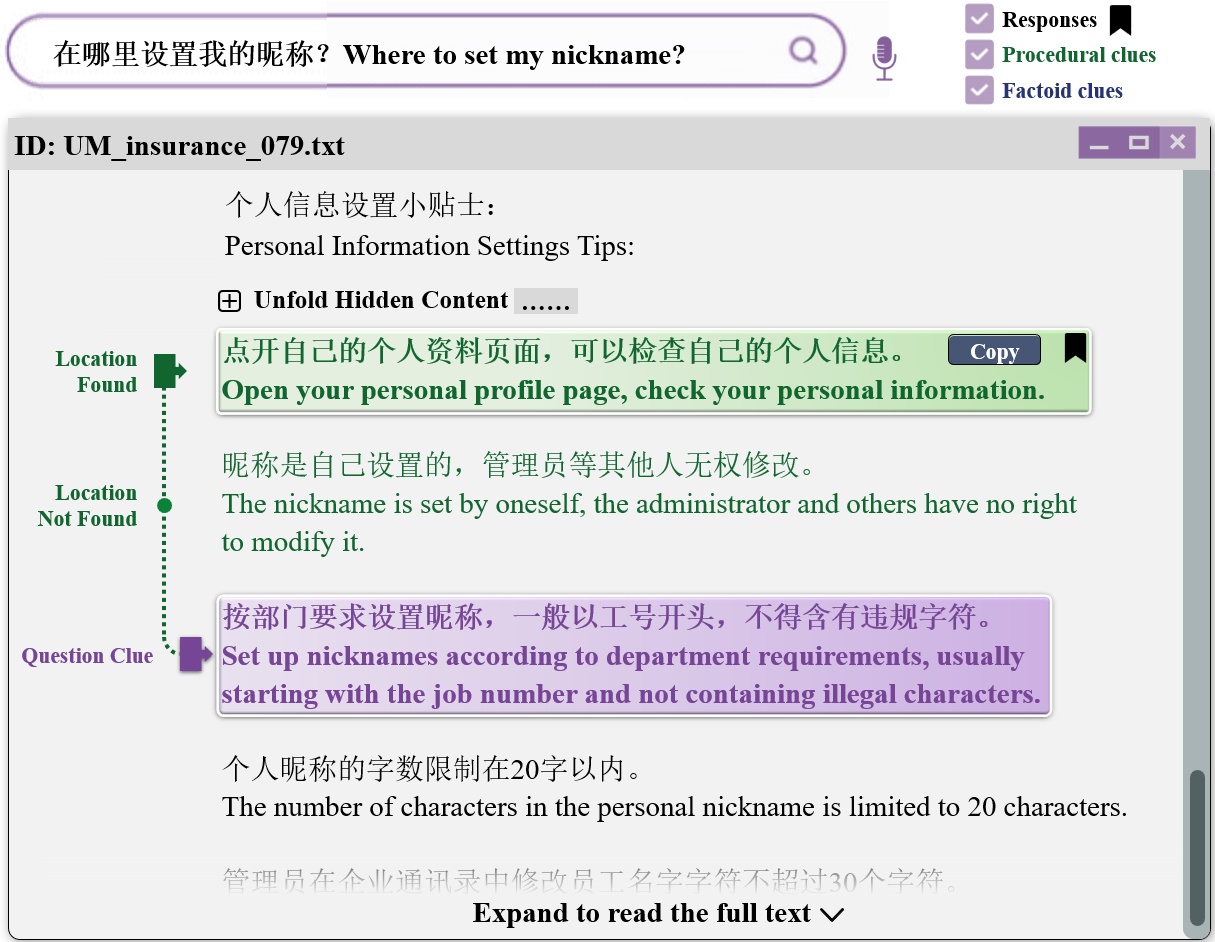}\label{fig:case1.3}}
    
    \caption{An case of clues provided by different assistants for user question.}
    \label{fig:case1}
    \vspace*{-0cm}
\end{figure*}

\begin{figure*}[!th]
    \centering
    \subfigure[Co-reference Contents Highlighting]{\includegraphics[width=.45\textwidth]{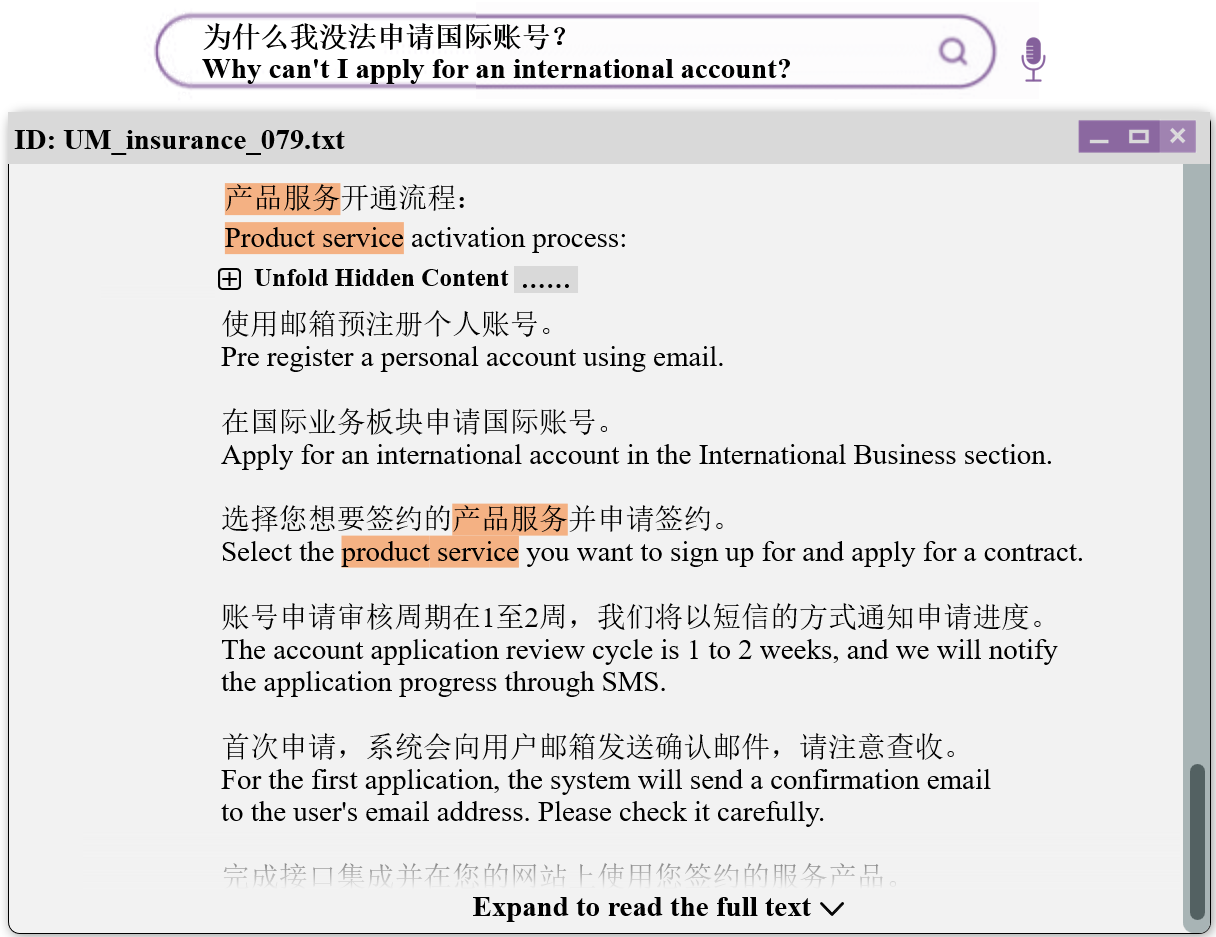}\label{fig:case2.4}}\hspace{1cm}
    \subfigure[Highlighting Specific Content]{\includegraphics[width=.45\textwidth]{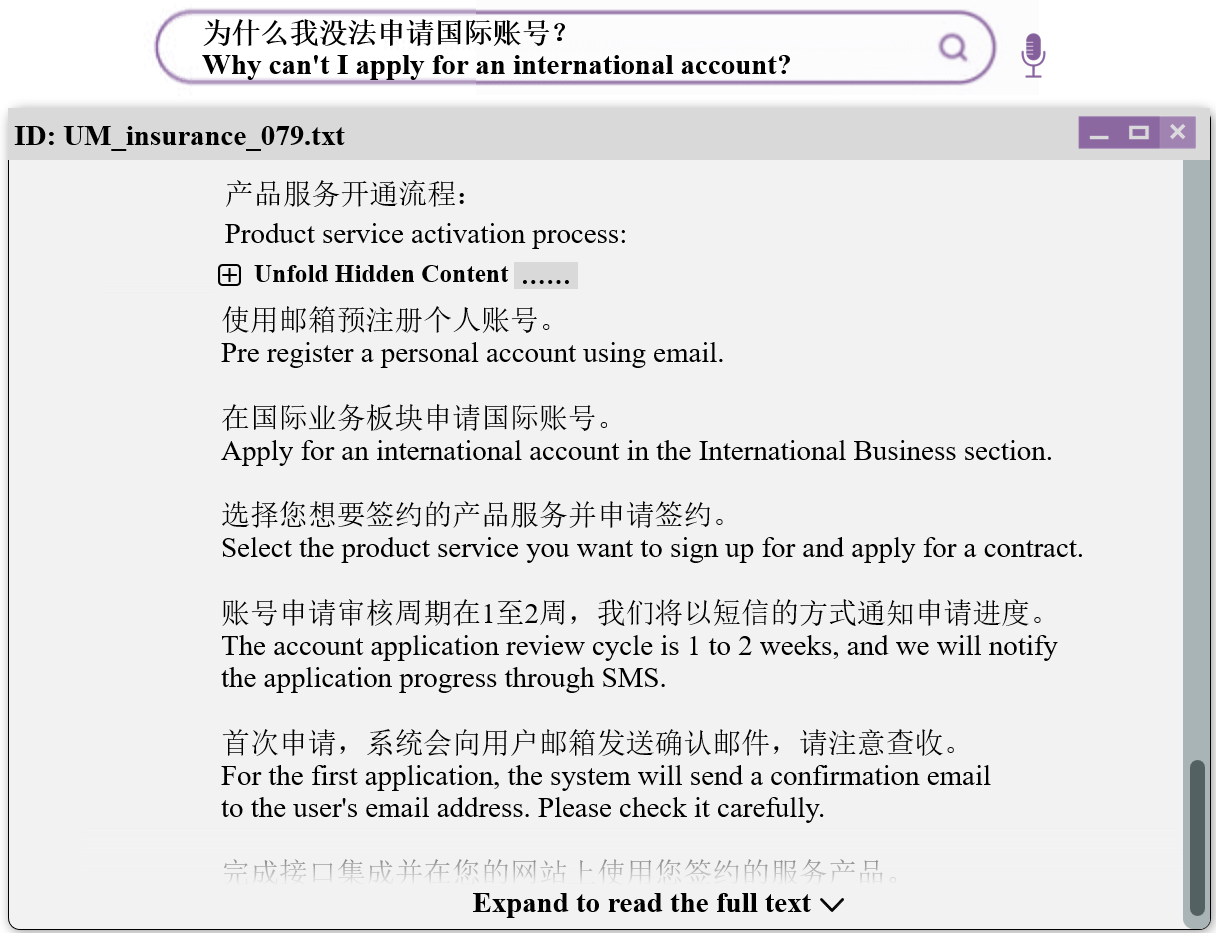}\label{fig:case2.5}}

    \vspace{1cm}
    \subfigure[Raw User Manual]{\includegraphics[width=.45\textwidth]{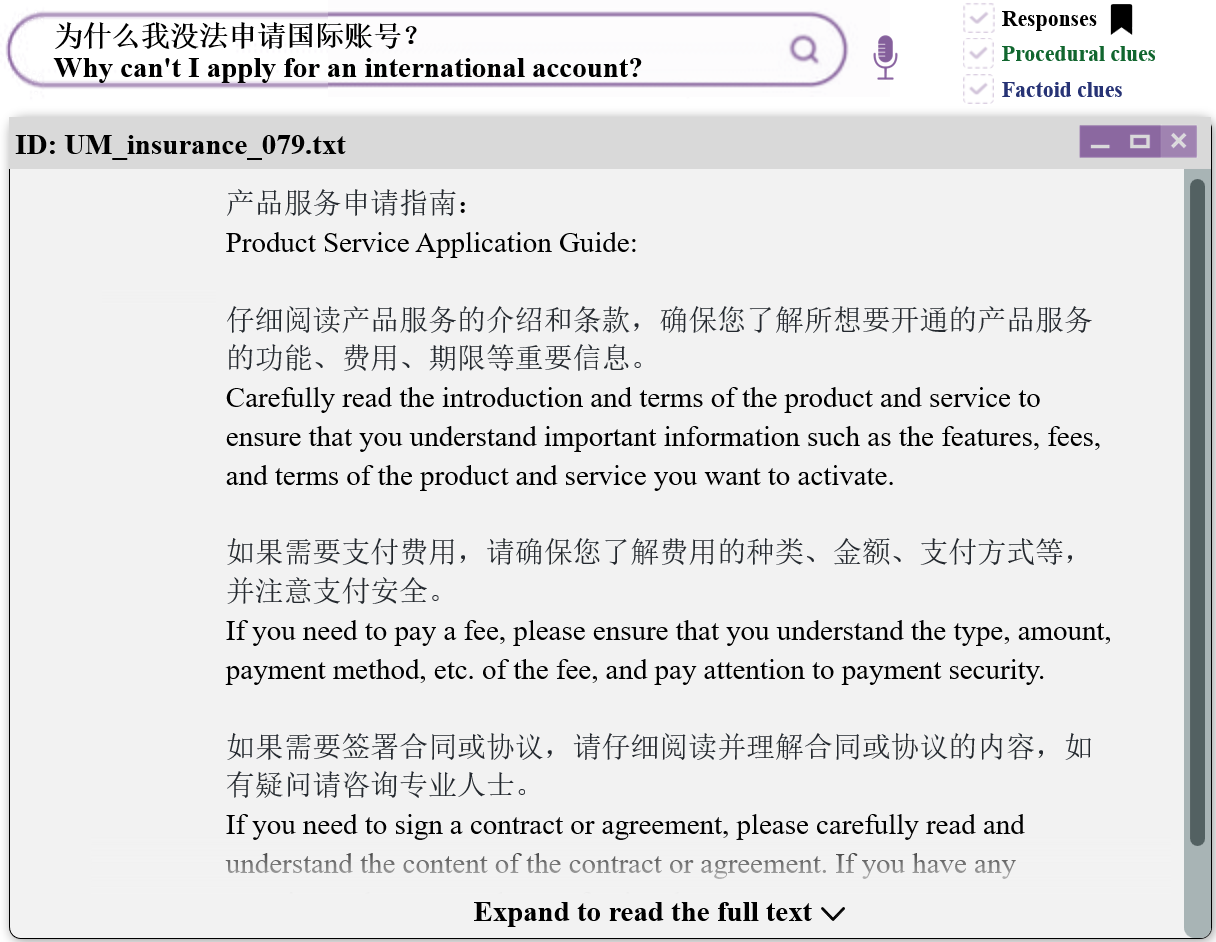}\label{fig:case2.1}}\hspace{1cm}
    \subfigure[Response Only]{\includegraphics[width=.45\textwidth]{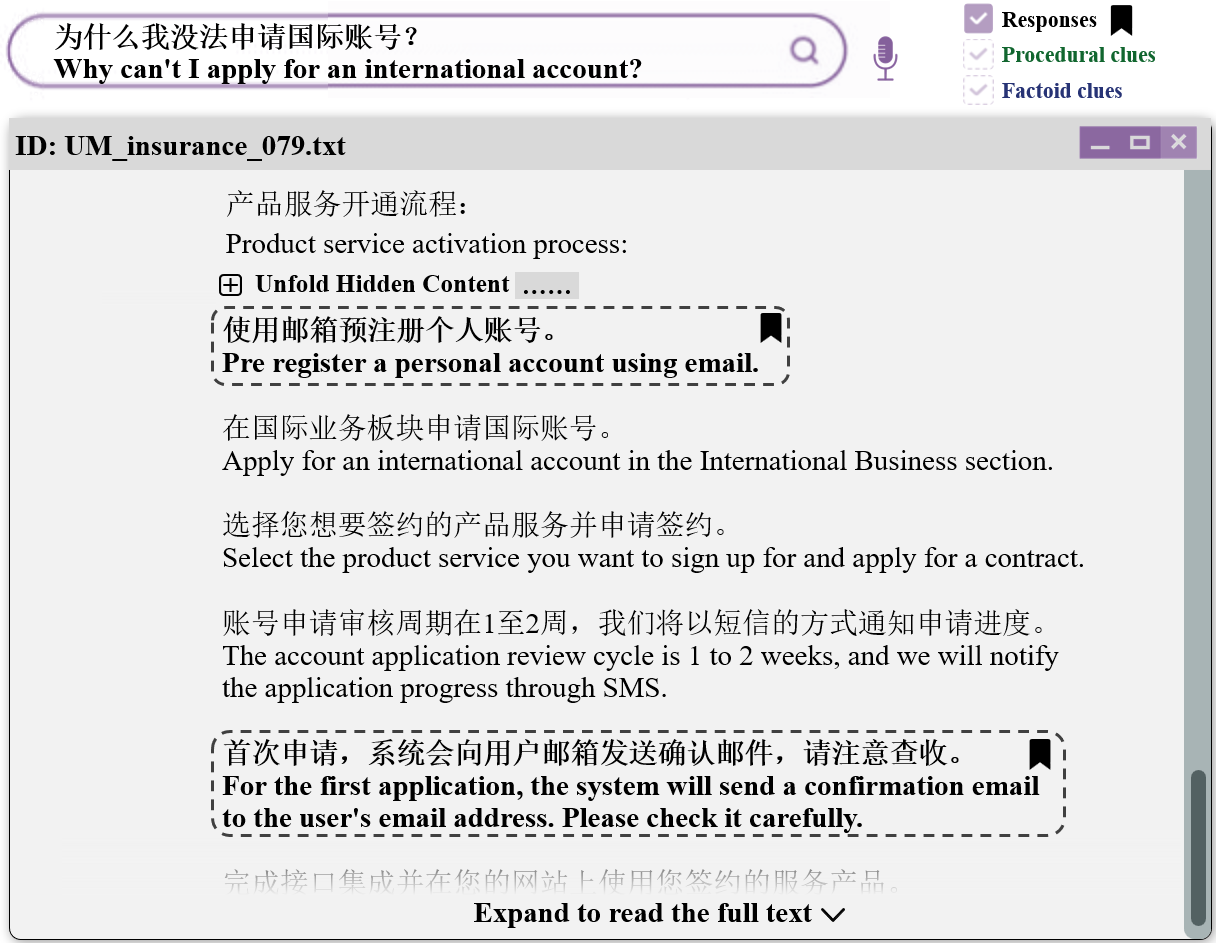}\label{fig:case2.2}}\hspace{1cm}

    \vspace{1cm}
    \subfigure[Clue-guided Assistant]{\includegraphics[width=.45\textwidth]{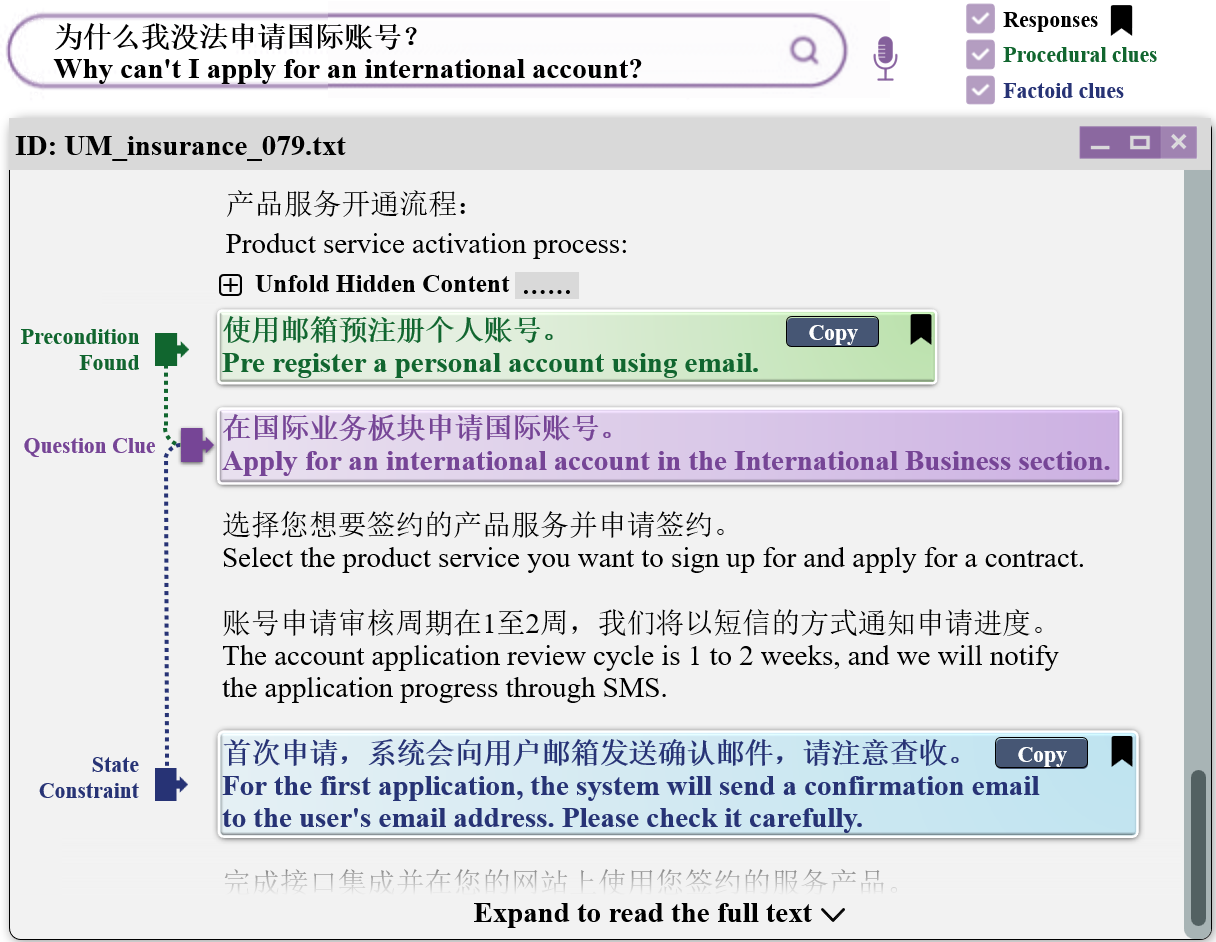}\label{fig:case2.3}}
    
    \caption{An case of clues provided by different assistants for inconsistent user questions.}
    \label{fig:case2}
    \vspace*{-0cm}
\end{figure*}
\section{Case Study}
Here we list some cases to demonstrate the superiority of our proposed assistant compared with other assistants. As shown in Figure~\ref{fig:case1}, for the question ``Where to set my nickname?'', the Co-reference Contents Highlighting approach~\ref{fig:case1.4} highlights all the spans ``nickname'' in the manual and the Marking Specific Content approach~\ref{fig:case1.5} highlights the ``kind tips'' section, both of them are nothing to do with the user question. Finally, our proposed assistant \sysname ~\ref{fig:case1.3} successfully locates the response ``personal profile page'' through in-depth inference on the procedural path. 
In more complex scenarios, as shown in Figure~\ref{fig:case2}, there is an inconsistency between the user question ``Why can't I apply for an international account?'' and corresponding manual, that is, specific action failure is not described directly in the user manual. The Co-reference Contents Highlighting approach~\ref{fig:case2.4} can still only highlight all the spans ``Product Service'' which are useless for answering the question, and the Marking Specific Content approach~\ref{fig:case2.5} even fails to match any spans. 
Unlike other assistants, our \sysname assistant~\ref{fig:case2.3} can conduct joint reference on both procedural and factual paths and successfully infers the procedural clue and factual clue for the failure of action ``apply for an international account'' mentioned in the question. This demonstrates that our proposed assistant can conduct in-depth inference on constructed graphs of manuals and is able to sufficiently locate potential responses through joint inference, further supporting CSRs in handling more complex user questions. 
% \clearpage

\end{document}